\PassOptionsToPackage{protrusion=false}{microtype}
\documentclass[11pt]{bluue}
\usepackage[T1]{fontenc}
\usepackage[utf8]{inputenc}
\usepackage{amsmath,amssymb,graphicx,booktabs,multirow,tabularx,microtype,xcolor}
\usepackage{colortbl,array,algpseudocode,fvextra}
\usepackage{placeins,enumitem,needspace}
\usepackage[most]{tcolorbox}
\usepackage{xspace}
\usepackage{xurl,hyperref}
\usepackage{eso-pic} 
\definecolor{vcgreen}{HTML}{557B48}
\definecolor{vcmint}{HTML}{EDF4E8}
\definecolor{vcblue}{HTML}{4E7DAC}
\definecolor{vclavender}{HTML}{ECE8F5}
\definecolor{vcpurple}{HTML}{8C75B8}
\definecolor{vcorange}{HTML}{A96D3A}
\definecolor{vchead}{HTML}{F2F5F7}
\definecolor{vcmultihead}{HTML}{E8F2F3}
\definecolor{vcsinglehead}{HTML}{EFEBF6}
\definecolor{vcbaseline}{HTML}{F3F4F6}
\definecolor{vcrestricted}{HTML}{F3F0F8}
\definecolor{vcjoint}{HTML}{DDEFE7}
\definecolor{vcdelta}{HTML}{FFF2D9}
\definecolor{vctransfera}{HTML}{EFF7F5}
\definecolor{vctransferb}{HTML}{F0F5FA}
\definecolor{vctransferc}{HTML}{F6F1F8}
\definecolor{rankbest}{HTML}{DCECF9}
\definecolor{ranksecond}{HTML}{F7E3ED}
\definecolor{rankworst}{HTML}{F2E8DC}
\newcommand{\best}[1]{\textbf{#1}}
\newcommand{\vcyes}{\raisebox{-1.2pt}{\includegraphics[height=8.7pt]{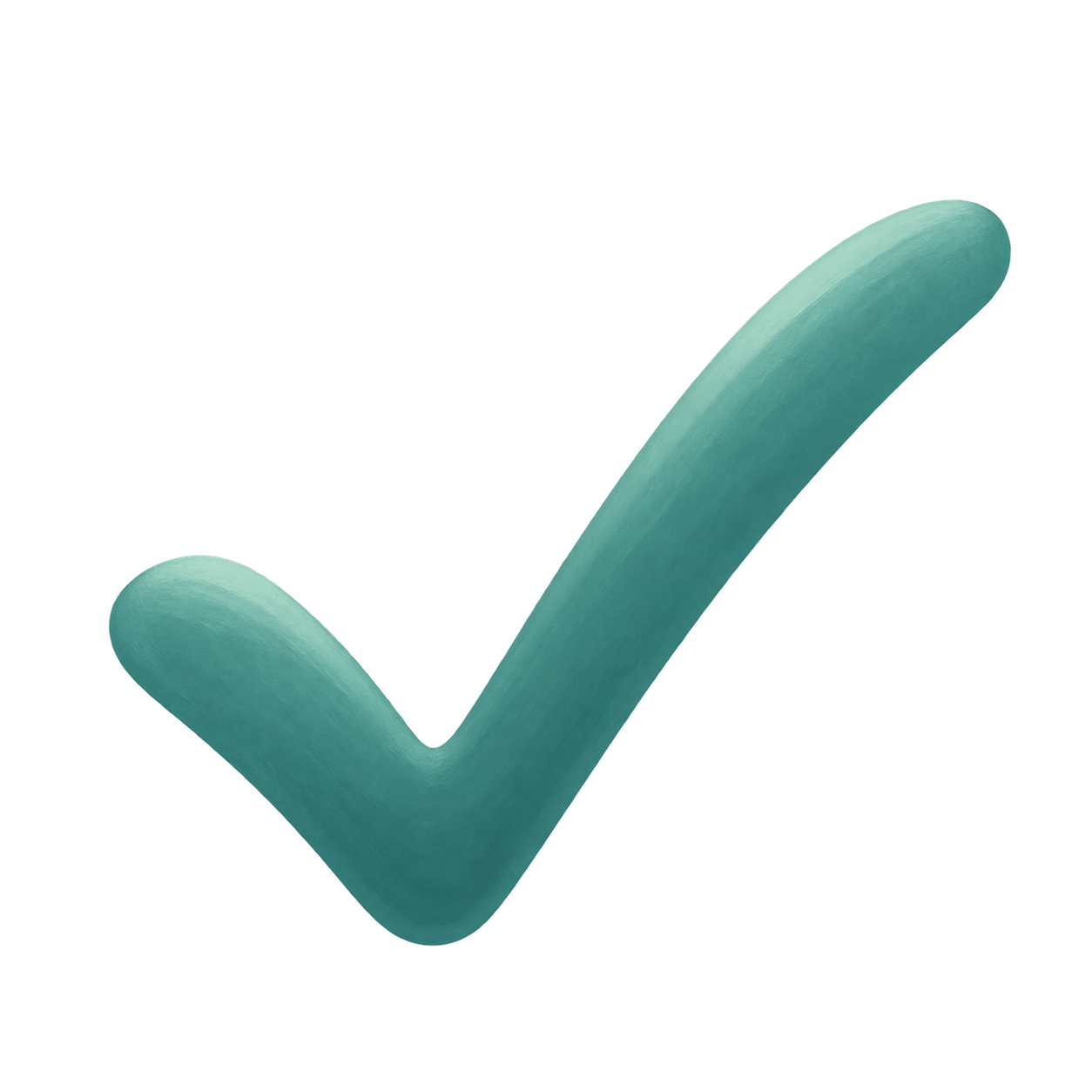}}}
\newcommand{\vcna}{\textcolor{gray}{\textendash}}
\newcommand{\vch}[1]{\begin{tabular}[c]{@{}c@{}}#1\end{tabular}}
\newcommand{\Wsp}{\mathcal{W}}
\hypersetup{pdftitle={ViewCraft: Evolving What Multimodal Agents See and How They Read It}}
\graphicspath{{./}{figures/}}
\renewcommand{\bottomfraction}{0}

\renewcommand{\floatpagefraction}{0.85}
\newcommand{\NAME}{\textsc{ViewCraft}\xspace}
\title{\textbf{\NAME: Evolving What Multimodal Agents See and How They Read It}}
\author[1]{\mbox{Binxiao Xu*}}
\author[1]{\mbox{Ruichuan An*}}
\author[3]{\mbox{Bocheng Zou}}
\author[5]{\mbox{Weihua Du}}
\author[6]{\mbox{Xinyu Wei}}
\author[1]{\mbox{Hangyu Guo}}
\author[4]{\mbox{YiFan Zhang}}
\author[1]{\mbox{Renye Yan}}
\author[7]{\mbox{Chengqi Duan}}
\author[2,\dagger]{\mbox{Hang Hua}}

\affiliation[1]{\mbox{PKU}}
\affiliation[2]{MIT-IBM Computing Research Lab}
\affiliation[3]{\mbox{UW--Madison}}
\affiliation[4]{\mbox{CASIA}}
\affiliation[5]{\mbox{CMU}}
\affiliation[6]{\mbox{PolyU}}
\affiliation[7]{\mbox{HKU}}

\DeclareUnicodeCharacter{2192}{\ensuremath{\rightarrow}}
\DeclareUnicodeCharacter{2014}{---}
\DeclareUnicodeCharacter{2013}{--}
\DeclareUnicodeCharacter{2019}{\textquotesingle}
\DeclareUnicodeCharacter{2018}{\textquotesingle}
\abstract{
Reusable skills improve multimodal agents without updating model weights,
but their effectiveness depends on both the evidence presented and the
procedure used to read it. We introduce ViewCraft, which jointly evolves
an executable evidence interface and its reading procedure. The interface
selects, transforms, and organizes observations; the reader interprets
the resulting views. Failure traces guide revisions, replay and paired
evaluation retain useful changes, and interface projection recovers
visual edits from eligible rejected joint proposals. Across eight
benchmarks, ViewCraft improves over direct execution by an average of
$6.1$ points in task score, with gains on every benchmark. The advantage
persists at a matched input-token budget. On all five multi-source tasks,
the learned interface uses $8$--$15\%$ fewer input tokens than Initial
skill. The learned skills also transfer across executors and outperform direct execution in every evaluated transfer, supporting evidence construction and reading as a coupled, reusable skill. Code is available on \href{https://github.com/Little-Fridge/ViewCraft}{GitHub}.
}

\makeatletter
\patchcmd{\mymaketitle}{\tcbset{top=6mm}}{\tcbset{top=3mm}}{}{\errmessage{Title top patch failed}}
\patchcmd{\mymaketitle}{\tcbset{bottom=6mm}}{\tcbset{bottom=3mm}}{}{\errmessage{Title bottom patch failed}}
\patchcmd{\mymaketitle}{\vspace*{0.65cm}}{}{}{\errmessage{Empty title field patch failed}}
\patchcmd{\mymaketitle}{\vskip 6mm}{\vskip 3mm}{}{\errmessage{Abstract spacing patch failed}}
\patchcmd{\maketitle}{\vskip 8mm}{\vskip 2mm}{}{\errmessage{Post-title spacing patch failed}}
\makeatother

\begin{document}
\AddToShipoutPictureFG*{%
  \AtTextLowerLeft{%
    \raisebox{-0.5\footskip}[0pt][0pt]{%
      \makebox[\textwidth][l]{%
        \normalfont\footnotesize\color{black}%
        \textsuperscript{*} Equal contribution.\quad
        \textsuperscript{\ensuremath{\dagger}} Corresponding author: Hang Hua
        (\href{mailto:hang.hua1@ibm.com}{hang.hua1@ibm.com}).%
      }%
    }%
  }%
}
\maketitle

\pagestyle{plain}
\begingroup
\setlength{\intextsep}{3pt}
\begin{figure}[H]
\centering
\includegraphics[width=0.9\linewidth]{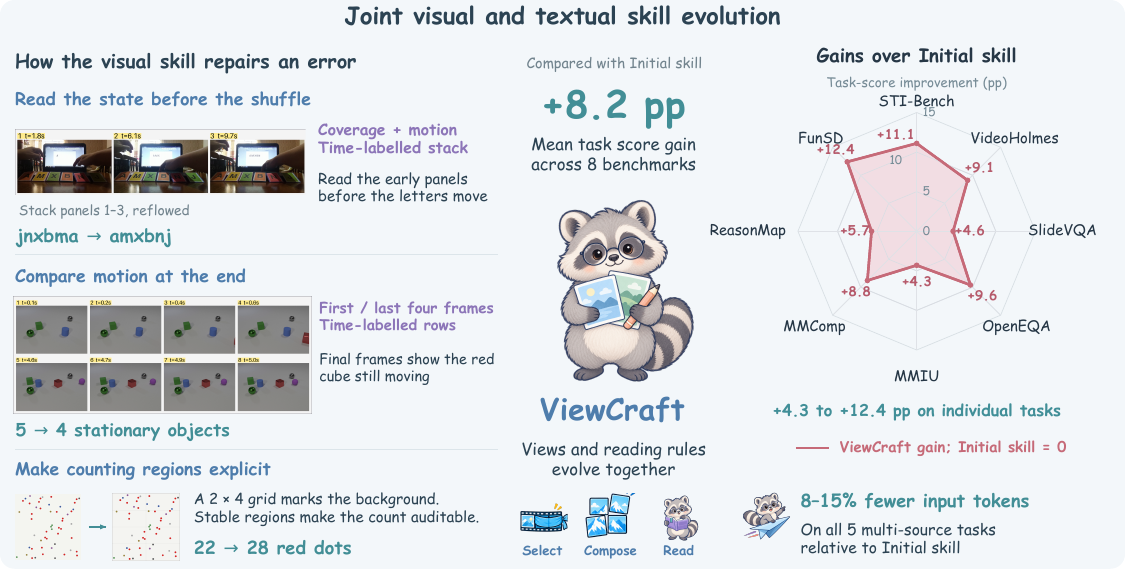}
\caption{\textbf{Joint evolution of evidence interfaces and reading procedures.} Left: early frames recover pre-shuffle letter order, endpoint rows expose motion, and a background grid supports counting. Center/right: gains over Initial skill use a shared $0$--$15$ point radial scale (mean $8.2$; range $4.3$--$12.4$ across eight tasks). On all five multi-source tasks, input tokens fall $8$--$15\%$ from Initial skill.}
\label{fig:motivation}
\end{figure}
\endgroup
\clearpage
\suppressfloats[t]
\setstretch{1.27}
\predisplaypenalty=10000
\brokenpenalty=10000
\begin{figure}[t]
\centering\includegraphics[width=\linewidth]{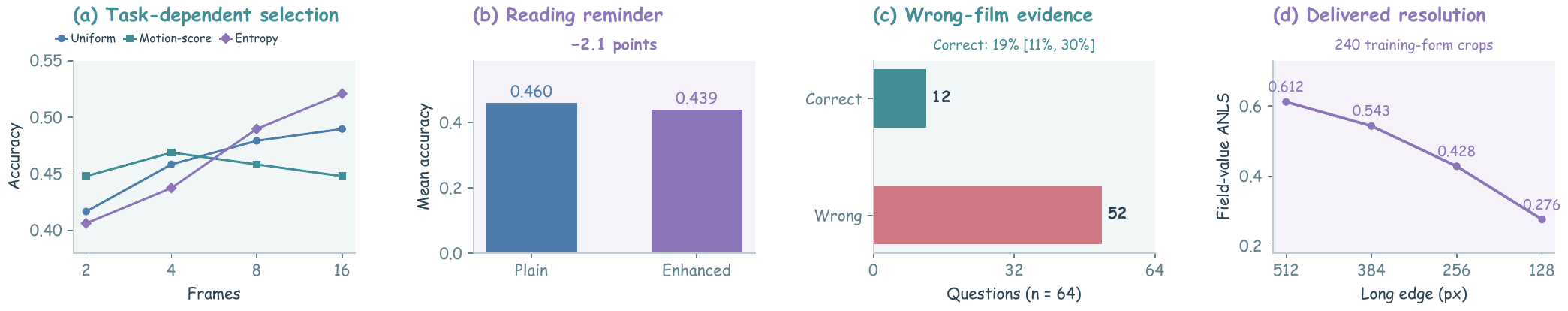}
\caption{\textbf{Four probes of the evidence interface.} (a) Accuracy for
three selectors at four frame budgets on 96 questions from 26 non-test
films. (b) Mean accuracy across these 12 settings with plain and enhanced
prompts. (c) With their frames replaced by another film, 12 of 64
previously correct questions are answered correctly (Wilson 95\% CI
$[11\%,30\%]$). (d) Mean field-value ANLS on 240 training-form crops at
four delivered resolutions. All probes are independent of
Table~\ref{tab:main}'s test sets.}
\label{fig:motiv-probes}
\vspace{-2mm}
\end{figure}
\begin{table}[t]
\caption{\textbf{What becomes part of an editable skill?} Columns track reuse, visual support, coupled editing, executable visual transformations, replay-screened paired evaluation, and interface-only retries. Checks denote explicit capabilities; a dash denotes a capability outside the described design. ViewCraft combines all six capabilities in its evolution loop.}
\label{tab:positioning}
\centering\footnotesize\setlength{\tabcolsep}{3pt}
\renewcommand{\arraystretch}{1.18}
\begin{tabularx}{\linewidth}{@{}X*{6}{c}@{}}
\toprule
\rowcolor{vchead}\textbf{Method family} & \vch{Reusable\\state} & \vch{Visual\\support} & \vch{Reader--visual\\co-edit} & \vch{Executable\\visual edits} & \vch{Replay +\\paired gate} & \vch{Interface\\retry}\\
\midrule
Textual skills & \vcyes & \vcna & \vcna & \vcna & \vcna & \vcna\\
Grounded multimodal skills & \vcyes & \vcyes & \vcna & \vcna & \vcna & \vcna\\
MPO~\citep{choi2026multimodal} & \vcyes & \vcyes & \vcyes & \vcna & \vcna & \vcna\\
Dynamo~\citep{sun2026dynamo} & \vcyes & \vcyes & \vcyes & \vcyes & \vcna & \vcna\\
\rowcolor{vcjoint}\textbf{\NAME} & \vcyes & \vcyes & \vcyes & \vcyes & \vcyes & \vcyes\\
\bottomrule
\end{tabularx}
\vspace{-2mm}
\end{table}
\begin{figure}[t]
\centering
\includegraphics[width=0.96\linewidth]{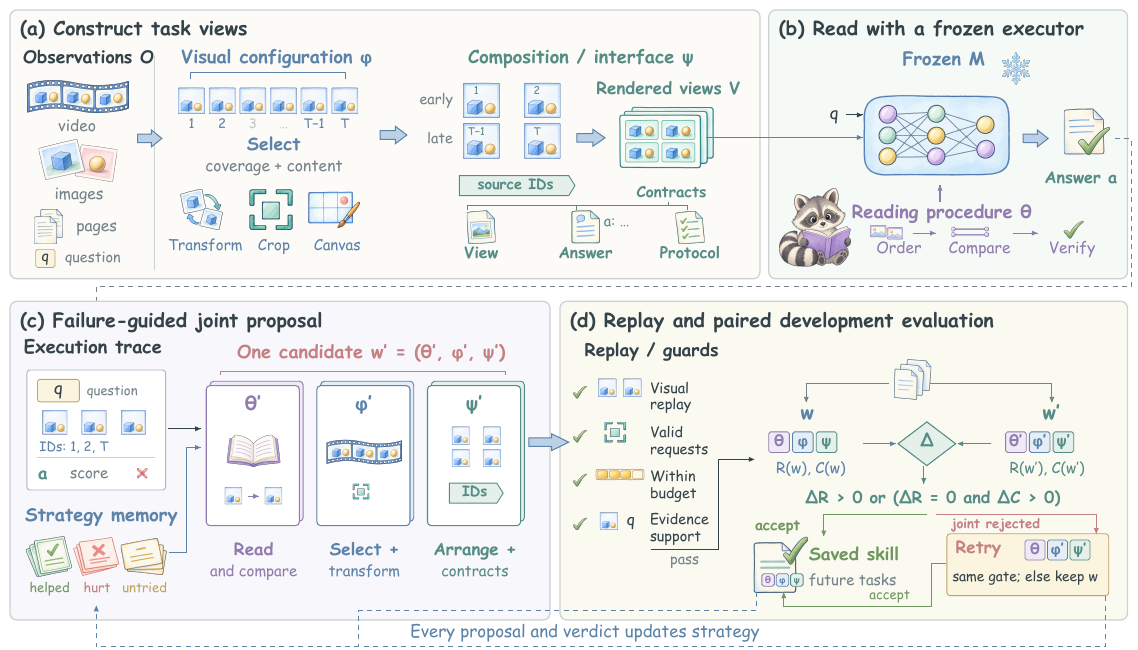}
\caption{\textbf{Joint view construction and evolution.} (a) $\phi$ selects/transforms evidence; $\psi$ arranges views and surface contracts. (b) Frozen $M$ follows $\theta$. (c) Traces guide proposals. (d) Replay and guards precede the paired gate. Projection retries the interface with incumbent $\theta$ when support is non-degraded and budget remains. Every verdict updates strategy.}
\label{fig:method}

\end{figure}
\begin{table}[t]
\caption{\textbf{Held-out task scores and component contributions.} Scores pool three runs under Table~\ref{tab:splits}'s metrics. The Mean column averages tasks equally; the final row subtracts Direct. Bold marks column bests. Tints distinguish baselines, restored-coordinate states, and joint evolution; lavender and teal headers group single-source and multi-source tasks.}
\label{tab:main}
\centering\small
\renewcommand{\arraystretch}{1.18}
\setlength{\tabcolsep}{4.2pt}
\resizebox{\textwidth}{!}{%
\begin{tabular}{@{}lrrrrrrrrr@{}}
\toprule
\rowcolor{vchead}\textbf{Method} & \cellcolor{vcsinglehead}\textbf{MMComp} & \cellcolor{vcsinglehead}\textbf{ReasonMap} & \cellcolor{vcsinglehead}\textbf{FunSD} & \cellcolor{vcmultihead}\textbf{STI-Bench} & \cellcolor{vcmultihead}\textbf{VideoHolmes} & \cellcolor{vcmultihead}\textbf{SlideVQA} & \cellcolor{vcmultihead}\textbf{OpenEQA} & \cellcolor{vcmultihead}\textbf{MMIU} & \textbf{Mean}\\
\midrule
\rowcolor{vcbaseline}Direct & 0.834 & 0.858 & 0.306 & 0.410 & 0.408 & 0.802 & 0.512 & 0.921 & 0.631\\
\rowcolor{vcbaseline}Initial skill & 0.786 & 0.841 & 0.288 & 0.385 & 0.381 & 0.812 & 0.478 & 0.908 & 0.610\\
Text-only & 0.848 & 0.876 & 0.372 & 0.430 & 0.424 & 0.837 & 0.528 & 0.926 & 0.655\\
Interface-only & 0.779 & 0.856 & 0.341 & 0.466 & 0.452 & 0.834 & 0.545 & 0.935 & 0.651\\
\rowcolor{vcrestricted}$\theta$-only & 0.852 & 0.867 & 0.352 & 0.416 & 0.408 & 0.833 & 0.508 & 0.921 & 0.645\\
\rowcolor{vcrestricted}$(\phi,\psi)$-only & 0.784 & 0.850 & 0.325 & 0.448 & 0.430 & 0.828 & 0.526 & 0.926 & 0.640\\
\rowcolor{vcjoint}\textbf{ViewCraft} & \best{0.874} & \best{0.898} & \best{0.412} & \best{0.496} & \best{0.472} & \best{0.858} & \best{0.574} & \best{0.951} & \best{0.692}\\
\midrule
\rowcolor{vcdelta}\textbf{ViewCraft $-$ Direct} & +0.040 & +0.040 & +0.106 & +0.086 & +0.064 & +0.056 & +0.062 & +0.030 & +0.061\\
\bottomrule
\end{tabular}}
\end{table}
\begin{table}[t]
\caption{Search strategies with matched submissions: 250 per arm across 24 task--seed runs; each run has four rounds and 96 training rollouts. Conditions share executor and interface. Scores pool three runs; filtering determines candidate evaluations.}
\label{tab:strategies}
\centering\footnotesize
\setlength{\tabcolsep}{3pt}
\resizebox{\linewidth}{!}{%
\begin{tabular}{l cccccccc c}
\toprule
\rowcolor{vchead}Search strategy & MMComp & RMap & FunSD & STI & VHolm & Slide & OEQA & MMIU & Mean \\
\midrule
Direct (no search) & 0.834 & 0.858 & 0.306 & 0.410 & 0.408 & 0.802 & 0.512 & 0.921 & 0.631 \\
MPO-adapted (prompt space) & 0.852 & 0.869 & 0.368 & 0.437 & 0.428 & 0.834 & 0.524 & 0.922 & 0.654 \\
Dynamo-style (fixed tools) & 0.841 & 0.865 & 0.380 & 0.440 & 0.426 & 0.840 & 0.534 & 0.925 & 0.656 \\
Independent, then combine  & 0.857 & 0.878 & 0.378 & 0.468 & 0.452 & 0.842 & 0.551 & 0.933 & 0.670 \\
Staged (visual, then $\theta$) & 0.859 & 0.880 & 0.385 & 0.472 & 0.455 & 0.846 & 0.556 & 0.938 & 0.674 \\
Alternating ($\theta \leftrightarrow \phi,\psi$) & 0.861 & 0.885 & 0.391 & 0.478 & 0.461 & 0.849 & 0.562 & 0.940 & 0.678 \\
\rowcolor{vcjoint}\textbf{Joint (ViewCraft)} & \textbf{0.874} & \textbf{0.898} & \textbf{0.412} & \textbf{0.496} & \textbf{0.472} & \textbf{0.858} & \textbf{0.574} & \textbf{0.951} & \textbf{0.692} \\
\bottomrule
\end{tabular}}
\end{table}

\begin{figure}[t]
\centering
\includegraphics[width=\linewidth]{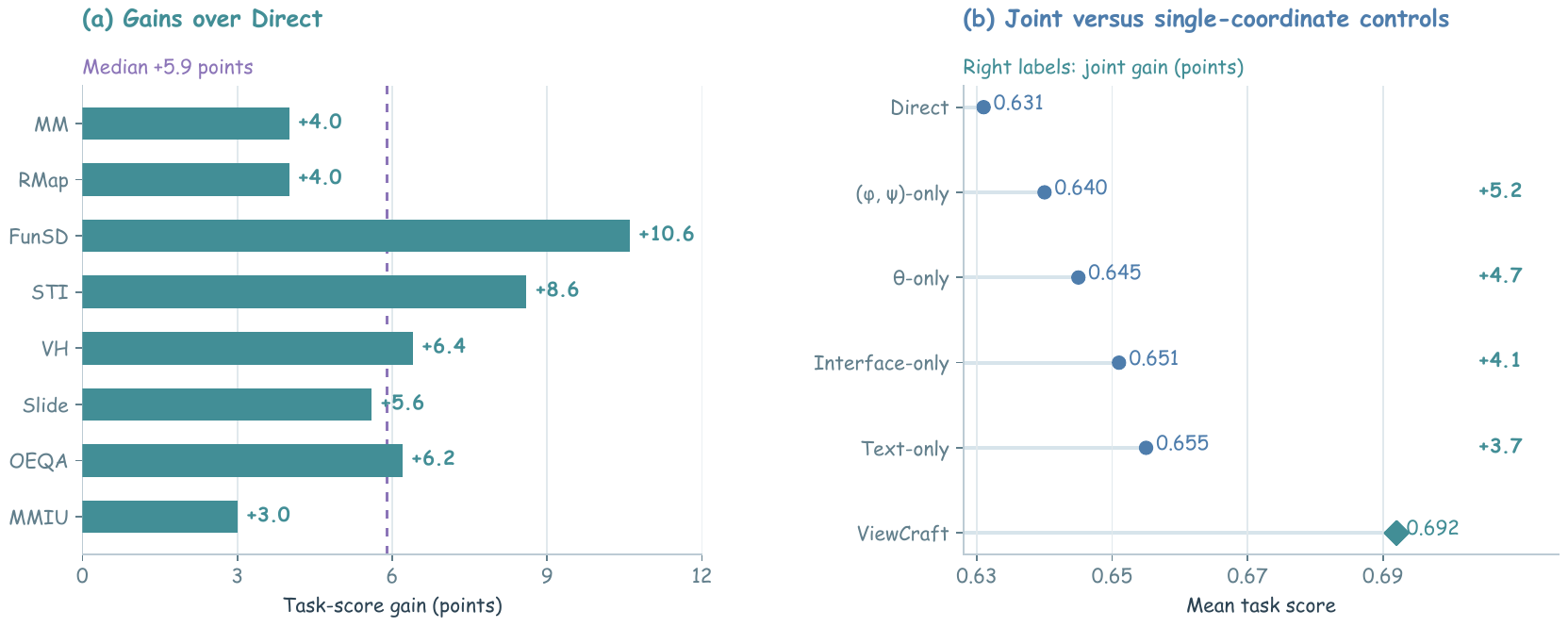}
\caption{\textbf{Where the gain sits.} (a) Per-benchmark held-out gains
over a bare Direct call (all eight positive; median $+5.9$ points).
(b) Mean task score of the four single-coordinate controls against joint
search (margins $+3.7$ to $+5.2$ points).}
\label{fig:gains}
\end{figure}
\begin{table}[t]
\caption{Baselines and value decomposition at matched budget
(per arm: 250 submitted proposals; per task--seed run: four rounds, 96
rollouts). Random proposals samples uniformly from the same expression
grammar under the same paired gate; MPO-fullspace and Dynamo-gen optimize
ViewCraft's executable interface; row 1 adapts prompt-space search; row 5
uses the full joint space and paired gate with replay, interface
projection, and the evidence tiebreak disabled.}
\label{tab:baselines}
\centering\footnotesize\setlength{\tabcolsep}{3pt}
\resizebox{\linewidth}{!}{%
\begin{tabular}{l cccccccc c}
\toprule
\rowcolor{vchead}& MMComp & RMap & FunSD & STI & VHolm & Slide & OEQA & MMIU & Mean\\
\midrule
MPO-adapted (prompt space) & 0.852 & 0.869 & 0.368 & 0.437 & 0.428 & 0.834 & 0.524 & 0.922 & 0.654\\
Random proposals (same space, gate) & 0.852 & 0.869 & 0.378 & 0.445 & 0.434 & 0.842 & 0.539 & 0.928 & 0.661\\
Dynamo-gen (tool generation) & 0.857 & 0.876 & 0.381 & 0.459 & 0.444 & 0.835 & 0.545 & 0.903 & 0.662\\
MPO-fullspace (UCB selection) & 0.861 & 0.878 & 0.385 & 0.463 & 0.450 & 0.838 & 0.549 & 0.907 & 0.666\\
Gate-only joint & 0.861 & 0.887 & 0.393 & 0.478 & 0.457 & 0.844 & 0.560 & 0.932 & 0.676\\
\rowcolor{vcjoint}\textbf{ViewCraft} & \textbf{0.874} & \textbf{0.898} & \textbf{0.412} & \textbf{0.496} & \textbf{0.472} & \textbf{0.858} & \textbf{0.574} & \textbf{0.951} & \textbf{0.692}\\
\bottomrule
\end{tabular}}
\vspace{-2mm}
\end{table}

\begin{figure}[t]
\centering
\includegraphics[width=\linewidth]{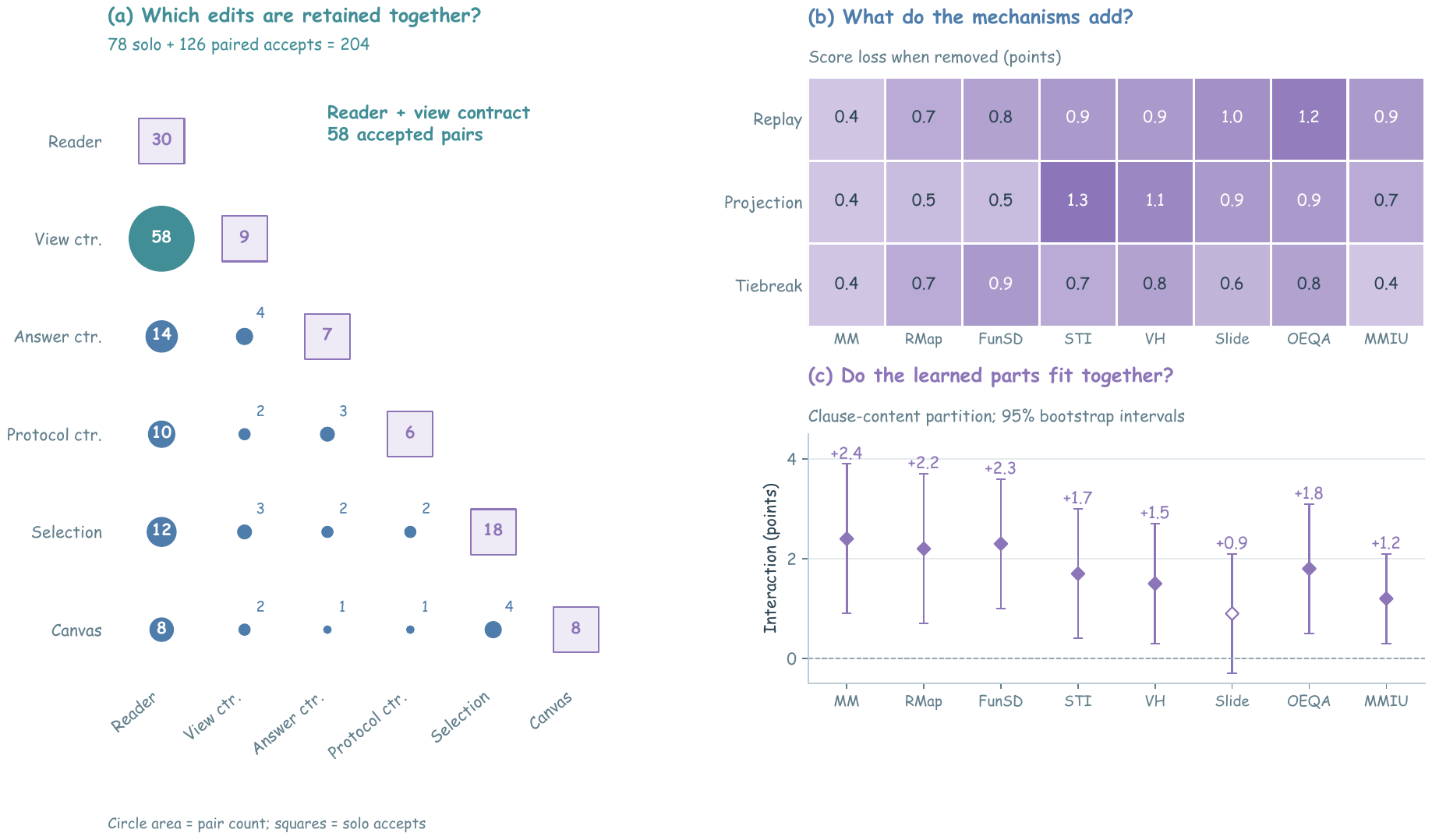}
\caption{\textbf{Matched search and co-adaptation.} (a) Accepted co-edits across the three-arm log: circle area encodes pair counts and diagonal squares solo accepts; total accepts 204. (b) Full-model score minus each ablation, in points (Table~\ref{tab:ablations}). (c) Terminal interaction $I$ under the clause-content partition; bars show 95\% seed-stratified media-cluster bootstrap intervals, including zero for SlideVQA.}
\label{fig:joint-necessity}\label{fig:coedits}
\vspace{-2mm}
\end{figure}

\begin{table}[t]
\caption{\textbf{Skill transfer across executors.} Bold marks the highest held-out score within each executor and task; tints group target executors. Self-transfer pools three runs; the Gemini diagonal reproduces Table~\ref{tab:main}. Cross-transfer uses the source's median-gate run. Direct uses the same pooling.}
\label{tab:transfer}
\centering\normalsize\renewcommand{\arraystretch}{2.0}\setlength{\tabcolsep}{2.8pt}
\begin{tabularx}{\linewidth}{@{}ll*{4}{>{\centering\arraybackslash}X}@{}}
\toprule
\rowcolor{vchead}\textbf{Executor} & \textbf{Skill source} & \cellcolor{vcsinglehead}\textbf{MMComp} & \cellcolor{vcsinglehead}\textbf{FunSD} & \cellcolor{vcmultihead}\textbf{STI-Bench} & \cellcolor{vcmultihead}\textbf{VideoHolmes}\\
\midrule
\rowcolor{vctransfera} & Direct & 0.834 & 0.306 & 0.410 & 0.408\\
\rowcolor{vctransfera} & gemini-3.8-flash & \textbf{0.874} & \textbf{0.412} & 0.496 & \textbf{0.472}\\
\rowcolor{vctransfera} & gpt-5.5 & 0.868 & 0.385 & \textbf{0.508} & 0.465\\
\rowcolor{vctransfera}\multirow{-4}{*}{gemini-3.8-flash} & claude-sonnet-4-6 & 0.852 & 0.379 & 0.484 & 0.458\\
\midrule
\rowcolor{vctransferb} & Direct & 0.850 & 0.271 & 0.567 & 0.550\\
\rowcolor{vctransferb} & gemini-3.8-flash & 0.871 & \textbf{0.388} & 0.623 & 0.598\\
\rowcolor{vctransferb} & gpt-5.5 & \textbf{0.881} & 0.377 & \textbf{0.641} & \textbf{0.612}\\
\rowcolor{vctransferb}\multirow{-4}{*}{gpt-5.5} & claude-sonnet-4-6 & 0.865 & 0.354 & 0.627 & 0.593\\
\midrule
\rowcolor{vctransferc} & Direct & 0.623 & 0.221 & 0.520 & 0.444\\
\rowcolor{vctransferc} & gemini-3.8-flash & 0.693 & 0.308 & 0.556 & 0.495\\
\rowcolor{vctransferc} & gpt-5.5 & 0.702 & 0.298 & 0.564 & \textbf{0.518}\\
\rowcolor{vctransferc}\multirow{-4}{*}{claude-sonnet-4-6} & claude-sonnet-4-6 & \textbf{0.709} & \textbf{0.318} & \textbf{0.569} & 0.507\\
\bottomrule\end{tabularx}
\end{table}
\begin{figure}[t]
\centering
\includegraphics[width=\linewidth]{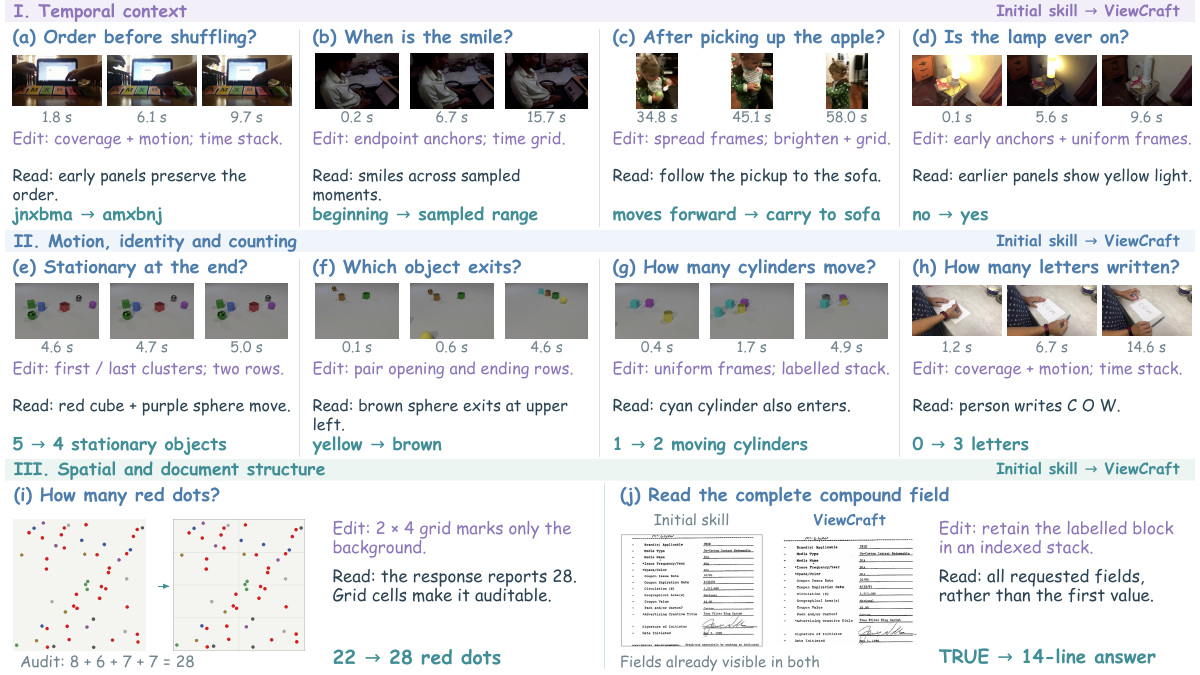}
\caption{\textbf{Ten repairs linking views to readings.} Panels link questions, view edits, reading summaries, and answer changes.}
\label{fig:cases-overview}
\vspace{-2mm}
\end{figure}

\Needspace{5\baselineskip}
\section{Introduction}
Multimodal reasoning needs relevant, legible evidence and a procedure matched to its organization. A missed video segment or cropped-away field obstructs downstream reasoning; a reader written for the original input may also overlook timestamps or spatial relations introduced by a new view. We ask whether the interface exposes the needed evidence and whether its reader uses that organization.

An end-state counting question illustrates both requirements. Early and late frames establish which objects moved; timestamped rows expose that comparison. The reader must match objects across rows before counting those that remained stationary. Figure~\ref{fig:motivation} shows temporal and counting repairs.

Skill evolution improves frozen executors through reusable procedures and workflows~\citep{yang2024large,khattab2024dspy,fernando2024promptbreeder,hua2025v2xum,hua2024finematch,zhang2025aflow,hu2025automated,yang26}. Multimodal systems add visual experience, prompts, and tools~\citep{xu26,zhang26,jiang26,choi2026multimodal,sun2026dynamo,hua2025finecaption,hu2023promptcap}. We make the evidence interface a saved, executable mapping that selects, transforms, and composes observations for each question, enabling readers to replay the mapping's revisions and evaluate their effects.

\paragraph{Evidence-interface probes.}\label{sec:motiv-exp}
Neither a single selector nor more frames consistently wins: selector rankings reverse with frame budget, and Motion-score declines after its peak (Figure~\ref{fig:motiv-probes}). The probes use 96 VideoHolmes questions from 26 non-test films and 240 field crops from FunSD training forms; none of these probe samples overlaps with the main test sets.

A stronger reading reminder lowers mean accuracy on unchanged inputs. Reducing the resolution of oracle field crops also lowers field-value ANLS despite preserving the requested content. Selection and legibility therefore offer concrete targets for an editable evidence interface (Appendix~\ref{app:protocol}).

We introduce \textbf{\NAME}, which jointly evolves the evidence
presented to a frozen executor and the procedure used to read it.
The resulting skill preserves how observations are selected and organized, together with the procedure for interpreting their relationships.

Our contributions are threefold:
\begin{itemize}
   \item We formulate multimodal skill evolution as jointly refining what agents see and how they read it. The reusable skill pairs an executable evidence interface with its reader, incorporating instance-conditioned evidence selection, transformation, and presentation.

    \item We introduce \textbf{\NAME}, which uses failure traces, visual replay, and paired development evaluation to evolve the interface and reader while freezing the executor. Interface projection retests visual edits from eligible rejected joint proposals with the incumbent reader to retain useful interface changes together with the existing reading procedure.

    \item Across eight benchmarks, \NAME improves over Direct by 6.1 points on average and remains 5.2 points ahead at a matched input-token budget. It also outperforms restricted and same-space search baselines, and the learned skills transfer across executors.
\end{itemize}

\Needspace{4\baselineskip}
\section{Related Work}
\subsection{Feedback-driven agent optimization}
Automatic prompt optimizers generate, select, or evolve instructions
from task feedback
\citep{zhou2022large,yang2024large,fernando2024promptbreeder,hu2023promptcap,zeng2026epicon}, while DSPy
and GEPA optimize compound LM programs from execution trajectories
\citep{khattab2024dspy,agrawal2026gepa}. Language agents also retain
verbal feedback, distilled experience, or executable skills
\citep{shinn2023reflexion,zhao2024expel,wang2023voyager,tang2026caption}, and systems
such as ADAS, AFlow, and AgentSquare search broader workflows or module
configurations
\citep{hu2025automated,zhang2025aflow,shang2025agentsquare}. Most of this
state is textual or structural. ViewCraft makes the evidence interface
itself part of the persistent state and evolves it together with its
reader.
\subsection{Multimodal skill evolution}
Multimodal skills incorporate spatial priors and interleaved support~\citep{xu26}, state cards and keyframes~\citep{zhang26}, and grounded experience for planning and tool use~\citep{jiang26}. Visual prompting and adaptive selection expose relevant regions and moments~\citep{yang23,hu24,wang24a,wang25,yu25,zeng2026mementogui}. Skill optimizers refine reusable procedures through execution feedback~\citep{yang2024large,khattab2024dspy,yang26}, while WikiSkill accumulates knowledge across updates~\citep{tang26}. The Multimodal Prompt Optimizer (MPO) jointly optimizes textual and non-textual prompts~\citep{choi2026multimodal}, and Dynamo co-evolves reasoning skills and visual tools~\citep{sun2026dynamo}. Rejecting a joint proposal can also discard a visual revision that would remain useful with the incumbent reader.

\NAME saves an executable evidence interface together with its reader (Table~\ref{tab:positioning}). Replay reconstructs the views, and interface projection retests the visual component of eligible rejected joint proposals with the incumbent reader. Table~\ref{tab:baselines} compares this design with MPO and Dynamo adaptations operating in the same search space as ViewCraft.

\section{ViewCraft}\label{sec:method}
ViewCraft searches for a reusable reader--interface state while keeping the executor frozen. Paired evaluation retains revisions that improve the responses to development tasks (Figure~\ref{fig:method}).

\subsection{Skill state and objective}
For $q$, observations $O=(o_1,\ldots,o_n)$, and target $y$, skill $w=(\theta,\phi,\psi)$ selects evidence via $\phi$, arranges views and formats via $\psi$, and gives the frozen executor a reading procedure $\theta$.

The renderer $\Wsp_{\phi,\psi}$ constructs image attachments from observations and question context. For a frozen executor $M$, ViewCraft targets
\begin{equation}
\max_{\phi,\psi,\theta}\ \mathbb{E}_{(O,q,y)\sim\mathcal D}
\left[R\!\left(M\!\left(\Wsp_{\phi,\psi}(O,q),\theta\right),y\right)\right],\qquad M\ \text{frozen}.
\label{eq:objective}
\end{equation}
ViewCraft approaches Equation~\ref{eq:objective} through proposals, development evaluation, and rollback. Text-only edits $\theta$, Interface-only edits $(\phi,\psi)$, and joint search permits both.

\paragraph{Coordinate partition.} Reasoning and decision rules belong to $\theta$, including polarity, verdict tables, zoom decisions, and mark reading. Rendering, view metadata, and answer/action formats remain on the interface side. Appendix~\ref{app:details} details the partition used in terminal crossings.

\subsection{Complementary success factors}
Let $E$ mean that required evidence is selected and legible. Guessing or prior knowledge can also yield some correct answers without that evidence, so
\begin{equation}
\begin{aligned}
P(\mathrm{correct})&=\underbrace{P(E)}_{\alpha}\underbrace{P(\mathrm{correct}\mid E)}_{\beta}
 +\underbrace{P(\neg E)P(\mathrm{correct}\mid\neg E)}_{\gamma},\\
R^{*}(w)&=\alpha(\phi,\psi)\,\beta(\theta\mid\phi,\psi)\le P(\mathrm{correct}).
\end{aligned}
\label{eq:factor}
\end{equation}
Equation~\ref{eq:factor} separates evidence availability, successful reading, and correctness without required evidence. The wrong-film probe estimates correctness after evidence replacement as $0.19$ with interval $[0.11,0.30]$. The evidence-absence probability weights this conditional rate in $\gamma$.

Two controls test whether evidence delivery or input volume alone explains the gain.

Evolved views with the initial reader recover $+2.3$ of the $+8.2$-point joint gain under a fixed trajectory (Appendix Table~\ref{tab:fixedtraj}); token-matched Direct recovers $+0.9$ points (Appendix Table~\ref{tab:budget}). The remaining advantage connects evidence construction to its reading procedure.

\subsection{Failure-guided proposals}
Traces record the question, sent views with geometry and provenance, response, and score. They guide revisions such as selecting a missed event or cropping an unreadable field. Interface contracts describe the views, answer format, and available requests; the reader specifies how to use them.

External guards check execution budgets and valid requests ($C_{\mathrm{exec}}$), and support for question-derived evidence requirements ($C_{\mathrm{ground}}$). A frozen verifier checks question-derived requirements against the sent views, without reference answers (Appendix~\ref{app:mechanics}). Replay filters rendering edits in $\phi$ or $\psi$ that leave the views unchanged. Reader-only and answer-text-only edits bypass this pixel check.

\subsection{Paired evaluation}
Let $\Delta R$ be the paired change in task score on the development set and $\Delta C$ the change in evidence support. After execution checks pass, the gate accepts when
\begin{equation}
\text{accept}\iff \Delta R>0\quad\text{or}\quad(\Delta R=0\ \wedge\ \Delta C>0).
\label{eq:gate}
\end{equation}
After a joint candidate changes visual fields but loses task score ($\Delta R<0$), ViewCraft retests candidate $(\phi,\psi)$ with incumbent $\theta$ only if evidence support has not fallen ($\Delta C\ge0$) and submission and gate allowances remain. This \emph{interface projection} consumes another submission and paired evaluation (Algorithm~\ref{alg:round}). Acceptance advances $w$; rejection restores the incumbent.

At test time, the saved skill supplies views, actions, and reading rules without optimization. Strategy retains rejected proposals after rollback; only accepted expressions affect execution.

\Needspace{4\baselineskip}
\section{Experiments}
\label{sec:experiments}
\subsection{Setting and comparisons}

We evaluate eight visual benchmarks using frozen, source-disjoint splits:
MMComposition~\citep{hua2024mmcomposition},
ReasonMap~\citep{feng2026reasonmap},
FUNSD~\citep{jaume2019funsd},
STI-Bench~\citep{li2025sti},
Video-Holmes~\citep{cheng2025video},
SlideVQA~\citep{tanaka2023slidevqa},
OpenEQA~\citep{majumdar2024openeqa}, and
MMIU~\citep{meng2025mmiu}.
The first three concentrate evidence in a single image or form; the remaining
five require selecting and comparing observations distributed across sources.

\paragraph{Score accounting.} Main comparisons pool three repetitions: each independently evolved terminal skill is tested once, while Direct and Initial skill are each executed three times. Binary scores use $S/(3n)$; continuous scores average the same repeated evaluations. FunSD uses field-value ANLS. Appendix~\ref{app:protocol} specifies construction, execution, and scoring.

Direct uses a bare call; Initial skill uses the empty-start interface. Text-only and Interface-only independently search the two sides under the clean partition. Terminal $\theta$-only and $(\phi,\psi)$-only restore one side of the joint result to its initial state, testing dependence within the learned pair.

Training traces guide proposals; gate items determine acceptance. Test questions remain outside both stages, and no form, film, or deck crosses the frozen splits.

Each evolving method runs 24 task--seed searches under shared proposal and rollout budgets.

The training rollouts for each search follow the allocation specified in Appendix~\ref{app:protocol}. Every search has four rounds and permits up to four submissions per round.

Gate sets range from 48 to 240 paired items (Table~\ref{tab:splits}); candidate and incumbent use the same items. Halving or doubling these sets changes mean task score by at most $0.2$ points (Appendix Table~\ref{tab:gatesens}).

\paragraph{Adapted baselines.} MPO-adapted retains paired text/non-text updates; its image-side prompt controls frame selection, crops, and grid or stack attachments. Dynamo-style composes reusable select-frame, crop-box, and tile-grid calls, with rendering primitives held as library code. These implementations share the executor, splits, paired gate, and submission budget.

Across its runs, each arm submits 250 candidates and performs 200--208 paired evaluations. Retries consume submissions; replay filters unchanged renderings. With submission counts matched, the mechanisms still determine which candidates reach the evaluation gate.

\Needspace{4\baselineskip}
\subsection{Task scores and coordinate effects}
Joint evolution leads every benchmark (Table~\ref{tab:main}; Figure~\ref{fig:gains}). Reading changes improve use of visible evidence, while interface edits expose relations lost through sampling or layout. Their relative value varies by task: SlideVQA benefits more from reading changes despite requiring multiple sources. Joint search accommodates bottlenecks in both reading and evidence presentation.

The terminal crossing writes $v=(\phi,\psi)$ and measures
\begin{equation}
I=R(\theta_1,v_1)-R(\theta_1,v_0)-R(\theta_0,v_1)+R(\theta_0,v_0).
\label{eq:interaction}
\end{equation}
The learned reader and interface help each other: all terminal interactions are positive, with mean $+1.8$ points (one-sided sign test $p=0.004$). Seven bootstrap intervals exclude zero; SlideVQA's includes it (Figure~\ref{fig:joint-necessity}c). Timestamp rows and canvas marks make this dependence concrete: the interface exposes a relation that its reader knows how to use.

Co-adaptation remains positive on average under all three coordinate partitions (Appendix Table~\ref{tab:partition}). Moving render-referencing clauses changes the magnitude while preserving the direction under both clause-content and strict partitions. Each crossing measures dependence under its stated definition.

Every task-level estimate favors ViewCraft, with an exact sign-test $p=0.008$. Holm correction across the comparisons retains FunSD and STI-Bench as individually significant (Appendix Table~\ref{tab:uncertainty}). Tests average runs per item and resample whole source media; the appendix reports run variation.

\subsection{Search strategies at matched budget}
\label{sec:strategies}
Joint search leads alternating search by $+1.3$ points (cluster-bootstrap CI $[+0.6,+2.1]$). All eight task-level gaps are positive (exact sign-test $p=0.008$), as are the three seed means (Appendix~\ref{app:stats}).

The ladder in Table~\ref{tab:baselines} separates access to the executable interface, candidate selection, failure guidance, and retention. Random same-space proposals improve over prompt-space search; UCB selection and failure-guided proposals add successive gains. The complete retention pipeline then reaches the highest score. ViewCraft's mean task score exceeds that of the strongest same-space baseline by $+2.6$ points, with a confidence interval of $[+1.3,+3.9]$.

ViewCraft also leads Dynamo-gen and random proposals in the shared expression space. The ladder increments and corresponding paired statistics are reported in Appendix~\ref{app:stats}.

The ordering in Table~\ref{tab:strategies} reflects how components adapt to one another. Independent search combines separately learned partners; staged search fixes an interface before its reader adapts; alternating search revisits them in turn. A single joint proposal can revise both the relation exposed by the interface and the reading rule used to interpret that relation.

MPO-fullspace and Dynamo-gen give the adapted methods ViewCraft's executable search space.

Trace-guided proposals lead random same-space proposals by $+1.5$ points; the retention mechanisms add $+1.6$. Their common expression grammar makes these gains attributable to the evaluated search configurations rather than access to different edits.

An evaluation-matched joint arm retains a mean score of 0.691 versus 0.692 for full ViewCraft, preserving the advantage under both submission and evaluation matching (Appendix~\ref{app:protocol}).

\subsection{Mechanisms and evaluation cost}
The reader and view contract form the most frequently accepted pair, with 58 co-edits (Figure~\ref{fig:joint-necessity}a). Defining what panels contain frequently accompanies a change in how to read them. Solo accepts retain one-sided revisions; terminal crossings test clause-content co-adaptation.

\paragraph{Mechanism effects and their overlap.} Replay filters inert renderings, projection recovers useful interfaces from rejected joint proposals, and the evidence tiebreak resolves equal task scores. Removing a single stage costs $0.7$--$0.9$ points; removing all stages costs $1.6$ (Appendix Table~\ref{tab:ablations}). Every paired interval excludes zero. The joint loss reveals overlap among stages.

Asymmetric projection has the highest mean score among the retry designs (Appendix Table~\ref{tab:projdesign}). Keeping the verified interface while reverting the reader gains $+0.6$ points over the symmetric direction. Replay saves 30 joint-arm gate evaluations; all 21 projection retries enter the saved skill.

Removing the evidence tiebreak lowers held-out scores on every benchmark (Appendix Table~\ref{tab:ablations}). The verifier agrees with human consensus on 89\% of audited judgments (Appendix~\ref{app:mechanics}). 

Most of the improvement survives with answer and protocol contracts fixed: the contract-frozen condition remains within 0.5 points of full ViewCraft on average. Format-normalized rescoring also preserves the advantage on FunSD (Appendix Table~\ref{tab:evidence}).

The gate links local edits to retained skill quality by scoring complete responses on paired items.

Replay and guards save evaluations; projection preserves eligible visual improvements after joint rejection. Their ablations measure which improvements survive into deployment (Appendix~\ref{app:mechanics}).

\subsection{Transfer and input cost}
The learned pairing remains useful after changing the executor. On held-out questions, every transferred skill exceeds its target executor's Direct baseline by at least 1.5 points (Table~\ref{tab:transfer}).

Three transfers exceed target-evolved skills: GPT-5.5 improves Gemini on STI-Bench and Sonnet on VideoHolmes; Gemini improves GPT-5.5 on FunSD. The remaining transfers fall within 3.3 points of target-evolved skills, preserving the pairing's value across executors.

The MMComp instance illustrates what transfers (Appendix~\ref{app:skill}). Its reader grounds option nouns and builds a verdict table; the interface colors notes by option and explains their identities. These saved rules construct and interpret evidence for new questions without target-specific search.

Evidence construction trades input volume for relevant detail (Appendix Table~\ref{tab:token-counts}). It adds tokens over Direct while learned selection reduces them from Initial skill on multi-source tasks.

Token-matched Direct recovers $+0.9$ of the $+6.1$-point margin; three-sample self-consistency reaches a similar mean (Appendix Table~\ref{tab:budget}). The remaining gain connects evidence to its reader.

Reuse amortizes search, while execution adds inspection. Latency rises by 7.6 seconds per item over Direct (Appendix Table~\ref{tab:deploy}). Task score and execution cost characterize reuse.

\subsection{Case studies}
Figure~\ref{fig:cases-overview} shows how view edits expose usable relations. Endpoint frames reveal motion when the reader matches objects across rows; early letter panels recover the pre-shuffle state when read before later frames. Both repairs combine temporal coverage with a rule for interpreting it.

A background grid preserves dot colors while creating stable counting regions. The reader can localize partial counts and check them against the image, making existing evidence easier to use.

Both form inputs contain the requested fields, but the initial reader returns one value while the evolved reader preserves the complete compound query.

Responses must retain field--value associations and the required format. Appendix Figure~\ref{fig:cases-boundaries} shows regressions when field labels or required option letters are omitted.

\setstretch{1.20}
\Needspace{6\baselineskip}
\section{Conclusion}
ViewCraft makes evidence construction a persistent, editable part of a multimodal skill. The central design choice is to evolve what the executor sees together with how it reads. Frame selection determines which moments are available, layout exposes relationships across observations, and resolution controls whether fine details remain legible. A reading procedure turns these views into comparisons, counts, and decisions. The probes and recorded repairs connect these concrete operations to the two questions that motivate the work: whether the necessary evidence is visible, and whether the reader can use its spatial and temporal organization. Joint evolution improves both parts while keeping the executor frozen. It outperforms single-coordinate search and same-space adapted baselines, and its advantage over Direct persists at matched input budget. Replay and paired evaluation make candidate views inspectable and tie retention to task outcomes; interface projection recovers useful visual edits from rejected joint candidates. The resulting state can be applied to new questions and transferred across executors. On all five multi-source tasks, learned selection also sends $8$--$15\%$ fewer input tokens than Initial skill. These results identify an actionable unit of reuse: a pair consisting of an executable evidence interface and a reading procedure that uses the evidence rendered by that interface to answer new questions.

\label{main:end}
\setstretch{1}


\FloatBarrier
\bibliographystyle{unsrtnat}
\begingroup
\interlinepenalty=10000
\bibliography{vc_iclr2027_conference,multimodal_papers}

@article{zeng2026mementogui,
  title={MementoGUI: Learning Agentic Multimodal Memory Control for Long-Horizon GUI Agents},
  author={Zeng, Ziyun and Hua, Hang and Zou, Bocheng and Cai, Mu and Feris, Rogerio and Luo, Jiebo},
  journal={arXiv preprint arXiv:2605.18652},
  year={2026}
}

@inproceedings{hua2025finecaption,
  title={Finecaption: Compositional image captioning focusing on wherever you want at any granularity},
  author={Hua, Hang and Liu, Qing and Zhang, Lingzhi and Shi, Jing and Kim, Soo Ye and Zhang, Zhifei and Wang, Yilin and Zhang, Jianming and Lin, Zhe and Luo, Jiebo},
  booktitle={2025 IEEE/CVF Conference on Computer Vision and Pattern Recognition (CVPR)},
  pages={24763--24773},
  year={2025},
  organization={IEEE}
}

@inproceedings{hu2023promptcap,
  title={Promptcap: Prompt-guided image captioning for vqa with gpt-3},
  author={Hu, Yushi and Hua, Hang and Yang, Zhengyuan and Shi, Weijia and Smith, Noah A and Luo, Jiebo},
  booktitle={2023 IEEE/CVF International Conference on Computer Vision (ICCV)},
  pages={2951--2963},
  year={2023},
  organization={IEEE}
}

@inproceedings{hua2024finematch,
  title={Finematch: Aspect-based fine-grained image and text mismatch detection and correction},
  author={Hua, Hang and Shi, Jing and Kafle, Kushal and Jenni, Simon and Zhang, Daoan and Collomosse, John and Cohen, Scott and Luo, Jiebo},
  booktitle={European Conference on Computer Vision},
  pages={474--491},
  year={2024},
  organization={Springer}
}

@inproceedings{hua2025v2xum,
  title={V2xum-llm: Cross-modal video summarization with temporal prompt instruction tuning},
  author={Hua, Hang and Tang, Yunlong and Xu, Chenliang and Luo, Jiebo},
  booktitle={Proceedings of the AAAI Conference on Artificial Intelligence},
  volume={39},
  number={4},
  pages={3599--3607},
  year={2025}
}

@inproceedings{tang2026caption,
  title={Caption anything in video: Fine-grained object-centric captioning via spatiotemporal multimodal prompting},
  author={Tang, Yolo Yunlong and Bi, Jing and Huang, Chao and Liang, Susan and Shimada, Daiki and Hua, Hang and Xiao, Yunzhong and Song, Yizhi and Liu, Pinxin and Feng, Mingqian and others},
  booktitle={Proceedings of the AAAI Conference on Artificial Intelligence},
  volume={40},
  number={48},
  pages={41697--41699},
  year={2026}
}

@inproceedings{choi2026multimodal,
  author = {Yumin Choi and Dongki Kim and Jinheon Baek and Sung Ju Hwang},
  title = {{Multimodal Prompt Optimization: Why Not Leverage Multiple Modalities for MLLMs}},
  booktitle = {International Conference on Learning Representations},
  year = {2026},
  url = {https://proceedings.iclr.cc/paper_files/paper/2026/hash/6cb71a5c9585c9a9fdd246ae32b309d3-Abstract-Conference.html}
}

@inproceedings{hu24,
  title     = {Visual Sketchpad: Sketching as a Visual Chain of Thought for Multimodal Language Models},
  author    = {Hu, Yushi and Shi, Weijia and Fu, Xingyu and Roth, Dan and Ostendorf, Mari and Zettlemoyer, Luke and Smith, Noah A. and Krishna, Ranjay},
  booktitle = {Advances in Neural Information Processing Systems},
  year      = {2024},
  url       = {https://arxiv.org/abs/2406.09403}
}

@article{hua2024mmcomposition,
  author = {Hang Hua and Yunlong Tang and Ziyun Zeng and Liangliang Cao and Zhengyuan Yang and Hangfeng He and Chenliang Xu and Jiebo Luo},
  title = {{MMComposition: Revisiting the Compositionality of Pre-trained Vision-Language Models}},
  journal = {arXiv preprint arXiv:2410.09733},
  year = {2024},
  url = {https://arxiv.org/abs/2410.09733}
}

@article{jiang26,
  title = {{XSkill}: Continual Learning from Experience and Skills in Multimodal Agents},
  author = {Jiang, Guanyu and Su, Zhaochen and Qu, Xiaoye and Fung, Yi R.},
  journal = {arXiv preprint arXiv:2603.12056},
  year = {2026},
  eprint = {2603.12056},
  archivePrefix = {arXiv},
  primaryClass = {cs.AI},
  url = {https://arxiv.org/abs/2603.12056}
}

@article{sun2026dynamo,
  author = {Yutao Sun and Yanting Miao and Hao-Xuan Ma and others},
  title = {{Dynamo: Dynamic Skill-Tool Evolution for Vision-Language Agents}},
  journal = {arXiv preprint arXiv:2606.30185},
  year = {2026},
  url = {https://arxiv.org/abs/2606.30185}
}

@misc{tang26,
  author = {Liyan Tang and Cyrus Rashtchian and Chun-Sung Ferng and Andrew Tomkins and Da-Cheng Juan and Tu Vu},
  title = {{WikiSkill: Compiling agent experience into persistent knowledge for skill evolution}},
  year = {2026},
  howpublished = {arXiv preprint arXiv:2608.27454},
  url = {https://arxiv.org/abs/2608.27454}
}

@inproceedings{wang24a,
  title     = {{VideoAgent}: Long-Form Video Understanding with Large Language Model as Agent},
  author    = {Wang, Xiaohan and Zhang, Yuhui and Zohar, Orr and Yeung-Levy, Serena},
  booktitle = {Computer Vision -- ECCV 2024},
  pages     = {58--76},
  year      = {2024},
  doi       = {10.1007/978-3-031-72989-8_4},
  url       = {https://doi.org/10.1007/978-3-031-72989-8_4}
}

@inproceedings{wang25,
  title     = {{VideoTree}: Adaptive Tree-Based Video Representation for LLM Reasoning on Long Videos},
  author    = {Wang, Ziyang and Yu, Shoubin and Stengel-Eskin, Elias and Yoon, Jaehong and Cheng, Feng and Bertasius, Gedas and Bansal, Mohit},
  booktitle = {IEEE/CVF Conference on Computer Vision and Pattern Recognition},
  pages     = {3272--3283},
  year      = {2025},
  url       = {https://openaccess.thecvf.com/content/CVPR2025/html/Wang_VideoTree_Adaptive_Tree-based_Video_Representation_for_LLM_Reasoning_on_Long_CVPR_2025_paper.html}
}

@article{xu26,
 title={{Agent Skills Should Go Beyond Text: The Case for Visual Skills}},
 author={Xu, Binxiao and An, Ruichuan and Zou, Bocheng and Hua, Hang},
 journal={arXiv preprint arXiv:2606.01414}, year={2026},
 url={https://arxiv.org/abs/2606.01414v1}
}

@article{yang23,
  title   = {Set-of-Mark Prompting Unleashes Extraordinary Visual Grounding in {GPT-4V}},
  author  = {Yang, Jianwei and Zhang, Hao and Li, Feng and Zou, Xueyan and Li, Chunyuan and Gao, Jianfeng},
  journal = {arXiv preprint arXiv:2310.11441},
  year    = {2023},
  url     = {https://arxiv.org/abs/2310.11441}
}

@article{yang26,
  title   = {{SkillOpt}: Executive Strategy for Self-Evolving Agent Skills},
  author  = {Yang, Yifan and Gong, Ziyang and Huang, Weiquan and Yang, Qihao and Zhou, Ziwei and Huang, Zisu and Li, Yan and Gao, Xuemei and Dai, Qi and Liu, Bei and Qiu, Kai and Yang, Yuqing and Chen, Dongdong and Yang, Xue and Luo, Chong},
  journal = {arXiv preprint arXiv:2605.23904},
  year    = {2026},
  url     = {https://arxiv.org/abs/2605.23904}
}

@inproceedings{yu25,
  title     = {{Frame-Voyager}: Learning to Query Frames for Video Large Language Models},
  author    = {Yu, Sicheng and Jin, Chengkai and Wang, Huanyu and Chen, Zhenghao and Jin, Sheng and Zuo, Zhongrong and Xu, Xiaolei and Sun, Zhenbang and Zhang, Bingni and Wu, Jiawei and Zhang, Hao and Sun, Qianru},
  booktitle = {International Conference on Learning Representations},
  year      = {2025},
  url       = {https://arxiv.org/abs/2410.03226}
}

@article{zhang26,
  title = {{MMSkills}: Towards Multimodal Skills for General Visual Agents},
  author = {Zhang, Kangning and Shao, Shuai and Li, Qingyao and Lin, Jianghao and Fu, Lingyue and Wang, Shijian and Jiao, Wenxiang and Lu, Yuan and Liu, Weiwen and Zhang, Weinan and Yu, Yong},
  journal = {arXiv preprint arXiv:2605.13527},
  year = {2026},
  eprint = {2605.13527},
  archivePrefix = {arXiv},
  primaryClass = {cs.AI},
  url = {https://arxiv.org/abs/2605.13527}
}

@inproceedings{feng2026reasonmap,
  title={Reasonmap: Towards fine-grained visual reasoning from transit maps},
  author={Feng, Sicheng and Wang, Song and Ouyang, Shuyi and Kong, Lingdong and Song, Zikai and Zhu, Jianke and Wang, Huan and Wang, Xinchao},
  booktitle={Proceedings of the IEEE/CVF Conference on Computer Vision and Pattern Recognition},
  pages={41077--41088},
  year={2026}
}

@inproceedings{li2025sti,
  title={Sti-bench: Are mllms ready for precise spatial-temporal world understanding?},
  author={Li, Yun and Zhang, Yiming and Lin, Tao and Liu, XiangRui and Cai, Wenxiao and Liu, Zheng and Zhao, Bo},
  booktitle={2025 IEEE/CVF International Conference on Computer Vision (ICCV)},
  pages={5622--5632},
  year={2025},
  organization={IEEE}
}

@inproceedings{jaume2019funsd,
  title={Funsd: A dataset for form understanding in noisy scanned documents},
  author={Jaume, Guillaume and Ekenel, Hazim Kemal and Thiran, Jean-Philippe},
  booktitle={2019 international conference on document analysis and recognition workshops (ICDARW)},
  volume={2},
  pages={1--6},
  year={2019},
  organization={IEEE}
}

@article{zeng2026epicon,
  title={EpiCon: Collective Agent Learning through Co-Evolving Multimodal Memory},
  author={Zeng, Ziyun and Hua, Hang and Alshammari, Shaden and Feris, Rogerio and Freeman, William T and Luo, Jiebo},
  journal={arXiv preprint arXiv:2609.37923},
  year={2026}
}

@article{cheng2025video,
  title={Video-holmes: Can mllm think like holmes for complex video reasoning?},
  author={Cheng, Junhao and Ge, Yuying and Wang, Teng and Ge, Yixiao and Liao, Jing and Shan, Ying},
  journal={arXiv preprint arXiv:2505.21374},
  year={2025}
}

@inproceedings{tanaka2023slidevqa,
  title={Slidevqa: A dataset for document visual question answering on multiple images},
  author={Tanaka, Ryota and Nishida, Kyosuke and Nishida, Kosuke and Hasegawa, Taku and Saito, Itsumi and Saito, Kuniko},
  booktitle={Proceedings of the AAAI Conference on Artificial Intelligence},
  volume={37},
  number={11},
  pages={13636--13645},
  year={2023}
}

@inproceedings{majumdar2024openeqa,
  title={Openeqa: Embodied question answering in the era of foundation models},
  author={Majumdar, Arjun and Ajay, Anurag and Zhang, Xiaohan and Putta, Pranav and Yenamandra, Sriram and Henaff, Mikael and Silwal, Sneha and Mcvay, Paul and Maksymets, Oleksandr and Arnaud, Sergio and others},
  booktitle={2024 IEEE/CVF Conference on Computer Vision and Pattern Recognition (CVPR)},
  pages={16488--16498},
  year={2024},
  organization={IEEE}
}

@inproceedings{meng2025mmiu,
  title={Mmiu: Multimodal multi-image understanding for evaluating large vision-language models},
  author={Meng, Fanqing and Wang, Jin and Li, Chuanhao and Lu, Quanfeng and Tian, Hao and Yang, Tianshuo and Liao, Jiaqi and Zhu, Xizhou and Dai, Jifeng and Qiao, Yu and others},
  booktitle={International Conference on Learning Representations},
  volume={2025},
  pages={38405--38453},
  year={2025}
}

@inproceedings{agrawal2026gepa,
  title={Gepa: Reflective prompt evolution can outperform reinforcement learning},
  author={Agrawal, Lakshya A and Tan, Shangyin and Soylu, Dilara and Ziems, Noah and Khare, Rishi and Opsahl-Ong, Krista and Singhvi, Arnav and Shandilya, Herumb and Ryan, Michael J and Jiang, Meng and others},
  booktitle={International Conference on Learning Representations},
  volume={2026},
  pages={8479--8565},
  year={2026}
}

@article{shinn2023reflexion,
  title={Reflexion: Language agents with verbal reinforcement learning},
  author={Shinn, Noah and Cassano, Federico and Gopinath, Ashwin and Narasimhan, Karthik and Yao, Shunyu},
  journal={Advances in neural information processing systems},
  volume={36},
  pages={8634--8652},
  year={2023}
}

@inproceedings{zhao2024expel,
  title={Expel: Llm agents are experiential learners},
  author={Zhao, Andrew and Huang, Daniel and Xu, Quentin and Lin, Matthieu and Liu, Yong-Jin and Huang, Gao},
  booktitle={Proceedings of the AAAI Conference on Artificial Intelligence},
  volume={38},
  number={17},
  pages={19632--19642},
  year={2024}
}

@article{wang2023voyager,
  title={Voyager: An open-ended embodied agent with large language models},
  author={Wang, Guanzhi and Xie, Yuqi and Jiang, Yunfan and Mandlekar, Ajay and Xiao, Chaowei and Zhu, Yuke and Fan, Linxi and Anandkumar, Anima},
  journal={arXiv preprint arXiv:2305.16291},
  year={2023}
}

@inproceedings{
khattab2024dspy,
title={{DSP}y: Compiling Declarative Language Model Calls into State-of-the-Art Pipelines},
author={Omar Khattab and Arnav Singhvi and Paridhi Maheshwari and Zhiyuan Zhang and Keshav Santhanam and Sri Vardhamanan A and Saiful Haq and Ashutosh Sharma and Thomas T. Joshi and Hanna Moazam and Heather Miller and Matei Zaharia and Christopher Potts},
booktitle={The Twelfth International Conference on Learning Representations},
year={2024},
url={https://openreview.net/forum?id=sY5N0zY5Od}
}

@inproceedings{
zhou2022large,
title={Large Language Models are Human-Level Prompt Engineers},
author={Yongchao Zhou and Andrei Ioan Muresanu and Ziwen Han and Keiran Paster and Silviu Pitis and Harris Chan and Jimmy Ba},
booktitle={The Eleventh International Conference on Learning Representations },
year={2023},
url={https://openreview.net/forum?id=92gvk82DE-}
}

@inproceedings{yang2024large,
  title={Large language models as optimizers},
  author={Yang, Chengrun and Wang, Xuezhi and Lu, Yifeng and Liu, Hanxiao and Le, Quoc V and Zhou, Denny and Chen, Xinyun},
  booktitle={International Conference on Learning Representations},
  volume={2024},
  pages={12028--12068},
  year={2024}
}

@inproceedings{
hu2025automated,
title={Automated Design of Agentic Systems},
author={Shengran Hu and Cong Lu and Jeff Clune},
booktitle={The Thirteenth International Conference on Learning Representations},
year={2025},
url={https://openreview.net/forum?id=t9U3LW7JVX}
}

@inproceedings{
zhang2025aflow,
title={{AF}low: Automating Agentic Workflow Generation},
author={Jiayi Zhang and Jinyu Xiang and Zhaoyang Yu and Fengwei Teng and Xiong-Hui Chen and Jiaqi Chen and Mingchen Zhuge and Xin Cheng and Sirui Hong and Jinlin Wang and Bingnan Zheng and Bang Liu and Yuyu Luo and Chenglin Wu},
booktitle={The Thirteenth International Conference on Learning Representations},
year={2025},
url={https://openreview.net/forum?id=z5uVAKwmjf}
}

@inproceedings{shang2025agentsquare,
  title={Agentsquare: Automatic llm agent search in modular design space},
  author={Shang, Yu and Li, Yu and Zhao, Keyu and Ma, Likai and Liu, Jiahe and Xu, Fengli and Li, Yong},
  booktitle={International Conference on Learning Representations},
  volume={2025},
  pages={3841--3865},
  year={2025}
}

@inproceedings{zeng2026mira,
  title={Mira: Multimodal iterative reasoning agent for image editing},
  author={Zeng, Ziyun and Hua, Hang and Luo, Jiebo},
  booktitle={Proceedings of the IEEE/CVF Conference on Computer Vision and Pattern Recognition},
  pages={9563--9573},
  year={2026}
}

@article{yu2026aurora,
  title={Aurora: Unified Video Editing with a Tool-Using Agent},
  author={Yu, Yongsheng and Zeng, Ziyun and Xiao, Zhiyuan and Zhou, Zhenghong and Hua, Hang and Xiong, Wei and Luo, Jiebo},
  journal={arXiv preprint arXiv:2605.18748},
  year={2026}
}

@inproceedings{
fernando2024promptbreeder,
title={Promptbreeder: Self-Referential Self-Improvement via Prompt Evolution},
author={Chrisantha Fernando and Dylan Sunil Banarse and Henryk Michalewski and Simon Osindero and Tim Rockt{\"a}schel},
booktitle={Forty-first International Conference on Machine Learning},
year={2024},
url={https://openreview.net/forum?id=9ZxnPZGmPU}
}

@article{zheng2026pearl,
  title={{PEARL: Personalized Streaming Video Understanding Model}},
  author={Zheng, Yuanhong and An, Ruichuan and Lin, Xiaopeng and Liu, Yuxing and Yang, Sihan and Zhang, Huanyu and Li, Haodong and Zhang, Qintong and Zhang, Renrui and Li, Guopeng and Zhang, Yifan and Li, Yuheng and Zhang, Wentao},
  journal={arXiv preprint arXiv:2603.20422},
  year={2026},
  url={https://arxiv.org/abs/2603.20422}
}

@article{an2026genius,
  title={{GENIUS: Generative Fluid Intelligence Evaluation Suite}},
  author={An, Ruichuan and Yang, Sihan and Guo, Ziyu and Dai, Wei and Shen, Zijun and Li, Haodong and Zhang, Renrui and Wei, Xinyu and Li, Guopeng and Wu, Wenshan and Zhang, Wentao},
  journal={arXiv preprint arXiv:2602.11144},
  year={2026},
  url={https://arxiv.org/abs/2602.11144}
}

@article{an2025unictokens,
  title={{UniCTokens: Boosting Personalized Understanding and Generation via Unified Concept Tokens}},
  author={An, Ruichuan and Yang, Sihan and Zhang, Renrui and Shen, Zijun and Lu, Ming and Dai, Gaole and Liang, Hao and Guo, Ziyu and Yan, Shilin and Luo, Yulin and Zou, Bocheng and Yang, Chaoqun and Zhang, Wentao},
  journal={arXiv preprint arXiv:2505.14671},
  year={2025},
  url={https://arxiv.org/abs/2505.14671}
}

@misc{lin2025perceiveanythingrecognizeexplain,
  title={{Perceive Anything: Recognize, Explain, Caption, and Segment Anything in Images and Videos}},
  author={Weifeng Lin and Xinyu Wei and Ruichuan An and Tianhe Ren and Tingwei Chen and Renrui Zhang and Ziyu Guo and Wentao Zhang and Lei Zhang and Hongsheng Li},
  year={2025},
  eprint={2506.05302},
  archivePrefix={arXiv},
  primaryClass={cs.CV},
  url={https://arxiv.org/abs/2506.05302}
}

@article{an2024mc,
  title={{MC-LLaVA: Multi-Concept Personalized Vision-Language Model}},
  author={An, Ruichuan and Yang, Sihan and Zhang, Renrui and Lu, Ming and Jiang, Tianyi and Zeng, Kai and Luo, Yulin and Cao, Jiajun and Liang, Hao and Chen, Ying and She, Qi and Zhang, Shanghang and Zhang, Wentao},
  journal={arXiv preprint arXiv:2411.11706},
  year={2024},
  url={https://arxiv.org/abs/2411.11706}
}

@inproceedings{luo2024llm,
  title={{LLM as Dataset Analyst: Subpopulation Structure Discovery with Large Language Model}},
  author={Luo, Yulin and An, Ruichuan and Zou, Bocheng and Tang, Yiming and Liu, Jiaming and Zhang, Shanghang},
  booktitle={European Conference on Computer Vision},
  pages={235--252},
  year={2024},
  organization={Springer},
  doi={10.1007/978-3-031-73414-4_14},
  url={https://www.ecva.net/papers/eccv_2024/papers_ECCV/papers/04858.pdf}
}

@inproceedings{yan2026doless,
  author={Yan, Renye and Cheng, Jikang and Sun, Shikun and Sun, Yi and Wu, You and Peng, Wei and Wang, Zongwei and Liang, Ling and Xing, Junliang and Cai, Yimao},
  title={{Do Less, Achieve More: Do We Need Every-Step Optimization for RL Fine-tuning of Diffusion Models?}},
  booktitle={Proceedings of the IEEE/CVF Conference on Computer Vision and Pattern Recognition (CVPR)},
  year={2026},
  pages={16561--16571},
  url={https://openaccess.thecvf.com/content/CVPR2026/html/Yan_Do_Less_Achieve_More_Do_We_Need_Every-Step_Optimization_for_CVPR_2026_paper.html}
}

@inproceedings{yan2025entropyadaptive,
  author={Yan, RenYe and Cheng, Jikang and Gan, Yaozhong and Sun, Shikun and Wu, You and Yang, Yunfan and Ling, Liang and Lin, Jinlong and Zhu, Yeshuang and Zhou, Jie and Zhang, Jinchao and Xing, Junliang and Cai, Yimao and Huang, Ru},
  title={{Entropy-Adaptive Diffusion Policy Optimization with Dynamic Step Alignment}},
  booktitle={Proceedings of the IEEE/CVF International Conference on Computer Vision (ICCV)},
  year={2025},
  pages={1924--1934},
  url={https://openaccess.thecvf.com/content/ICCV2025/html/Yan_Entropy-Adaptive_Diffusion_Policy_Optimization_with_Dynamic_Step_Alignment_ICCV_2025_paper.html}
}

@article{yan2026pixelspace,
  title={{Pixel-Space Diffusion Transformers}},
  author={Renye Yan and Jikang Cheng and You Wu and Ling Liang and Wei Peng and Athanasios V. Vasilakos and Qingyu Zhao and Yu Zhang and Yimao Cai and Kilian M. Pohl and Guoying Zhao},
  journal={arXiv preprint arXiv:2607.17585},
  year={2026},
  url={https://arxiv.org/abs/2607.17585}
}

@article{feng2026m2a,
  title={{M2A: Multimodal Memory Agent with Dual-Layer Hybrid Memory for Long-Term Personalized Interactions}},
  author={Junyu Feng and Binxiao Xu and Jiayi Chen and Mengyu Dai and Cenyang Wu and Haodong Li and Bohan Zeng and Yunliu Xie and Hao Liang and Ming Lu and Wentao Zhang},
  journal={arXiv preprint arXiv:2602.07624},
  year={2026},
  url={https://arxiv.org/abs/2602.07624}
}

@article{yang2025smalllarge,
  title={{Small-Large Collaboration: Training-efficient Concept Personalization for Large VLM using a Meta Personalized Small VLM}},
  author={Sihan Yang and Huitong Ji and Shaolin Lu and Jiayi Chen and Binxiao Xu and Ming Lu and Yuanxing Zhang and Wenhui Dong and Wentao Zhang},
  journal={arXiv preprint arXiv:2508.07260},
  year={2025},
  url={https://arxiv.org/abs/2508.07260}
}

@article{xu2025jarvis,
  title={{Jarvis: Towards Personalized AI Assistant via Personal KV-Cache Retrieval}},
  author={Binxiao Xu and Junyu Feng and Shaolin Lu and Yulin Luo and Shilin Yan and Hao Liang and Ming Lu and Wentao Zhang},
  journal={arXiv preprint arXiv:2510.22765},
  year={2025},
  url={https://arxiv.org/abs/2510.22765}
}

@article{xu2026admir,
  title={{AD-MIR: Bridging the Gap from Perception to Persuasion in Advertising Video Understanding via Structured Reasoning}},
  author={Binxiao Xu and Junyu Feng and Xiaopeng Lin and Haodong Li and Zhiyuan Feng and Bohan Zeng and Shaolin Lu and Ming Lu and Qi She and Wentao Zhang},
  journal={arXiv preprint arXiv:2602.07625},
  year={2026},
  url={https://arxiv.org/abs/2602.07625}
}
\endgroup

\appendix
\predisplaypenalty=10000
\widowpenalty=10000
\clubpenalty=10000
\setcounter{topnumber}{3}\setcounter{dbltopnumber}{3}
\setcounter{bottomnumber}{2}\setcounter{totalnumber}{4}
\renewcommand{\bottomfraction}{0.8}
\renewcommand{\floatpagefraction}{0.7}
\makeatletter
\setlength{\@fpsep}{8pt plus 2pt minus 2pt}
\setlength{\@dblfpsep}{8pt plus 2pt minus 2pt}
\makeatother
\normalsize
\setstretch{1}
\FloatBarrier
\section{Evaluation protocol}\label{app:protocol}

All conditions use frozen splits, the same executor, and paired item identities. Direct uses one bare call; Initial skill uses the empty-start execution interface. Table~\ref{tab:splits} lists source units and scoring rules. Sampling uses the full test set below 500 questions, half at 500--1000, and a quarter above 1000, preserving whole source media and category coverage. The selection is frozen before runs.

\paragraph{Score accounting.} Every accuracy in the main comparisons is the
pooled mean over three repetitions with denominator $3n$: evolving
conditions are run as three independent searches (different seeds), each
final skill evaluated once on the held-out set; Direct and Initial skill are
executed three times each (repetition applies uniformly across conditions).
For binary metrics this makes every entry an integer multiple of $1/(3n)$.
FunSD, SlideVQA, and OpenEQA use continuous scores (field-value ANLS,
ANLS, and LLM-Match respectively), so their means are unconstrained. We obtain MMIU's pooled score of $685/720\approx0.951$ by aggregating the results of all three runs.

\begin{table}[htbp]
\caption{Evaluation units, source-disjoint splits, and scoring. Cells
report questions / unique source media. Train, gate, and test draws have
pairwise-disjoint source media; no form, film, diagram, or deck appears in
two splits. Gate questions are never shown to the proposal stage, which
uses training traces only.}
\label{tab:splits}
\centering\small
\setlength{\tabcolsep}{4pt}\renewcommand{\arraystretch}{1.05}
\resizebox{0.9\linewidth}{!}{\begin{tabular}{l l rrr l}
\toprule
\rowcolor{vchead}
Benchmark & Medium & Train & Gate & Test & Metric \\
\midrule
MMComp & image & 48 / 48 & 96 / 96 & 151 / 151 & option-letter accuracy \\
ReasonMap & transit diagram & 24 / 2 & 48 / 2 & 153 / 4 & route accuracy \\
FunSD & form & 64 / 58 & 96 / 72 & 248 / 47 & field-value ANLS \\
STI-Bench & video & 36 / 12 & 72 / 18 & 221 / 36 & option-letter accuracy \\
VideoHolmes & video & 24 / 10 & 72 / 16 & 183 / 28 & option-letter accuracy \\
SlideVQA & slide deck & 40 / 10 & 60 / 14 & 155 / 30 & ANLS \\
OpenEQA & video & 12 / 4 & 48 / 8 & 95 / 10 & LLM-Match (mean) \\
MMIU & image & 240 / 240 & 240 / 240 & 240 / 240 & option-letter accuracy \\
\bottomrule
\end{tabular}}
\end{table}
\begin{table}[htbp]
\caption{Gate-size sensitivity: mean task score when every benchmark's
gate set is scaled as shown, holding proposals, rollouts, and test sets
fixed.}
\label{tab:gatesens}
\centering\footnotesize
\begin{tabular}{l ccccc}
\toprule
\rowcolor{vchead}Gate scale (items/task) & 8 & 25\% & 50\% & 100\% (default) & 200\%\\
\midrule
Mean task score & 0.679 & 0.686 & 0.690 & 0.692 & 0.693\\
\bottomrule
\end{tabular}
\end{table}
\paragraph{Evidence-interface probes.}
The probes use a frozen executor on 96 VideoHolmes questions from 26 non-test films and 240 field crops from FunSD training forms. These samples have zero overlap with the main test sets. Three selectors use the same questions at four frame budgets; plain and enhanced prompts are compared over these settings. In the oracle-crop comparison, requested content is held fixed while the resolution of the images delivered to the executor varies.

\paragraph{Wrong-film probe.}
Replacing the frames for 64 previously correct questions with another film yields 12 correct and 52 wrong answers, with no abstentions or mismatch flags (Figure~\ref{fig:motiv-probes}c).

\paragraph{FunSD construction.}
FunSD is converted to field-value QA. Question-class entities become field labels; the reference answer is the vertically-nearest answer-class entity. Table~\ref{tab:splits} gives the constructed splits. The split unit is the form: no form appears in two
splits, so every field question derived from one form stays with it. The
metric is field-value ANLS (normalized edit similarity on the lowercased
transcription), continuous in $[0,1]$; Table~\ref{tab:main} reports the pooled mean.
A multi-line question entity forms a compound field whose reference retains the full multi-line value.

\paragraph{FunSD construction audit.}
Of 200 constructed pairs, 192 link the correct field (96.0\%); the eight errors come from multi-column forms (4), multi-line answers (3), and nearest-entity ambiguity (1). With gold labels, Direct shifts 0.306 to 0.312 and ViewCraft 0.412 to 0.415.

\begin{table}[htbp]
\caption{Per-arm development cost over 24 task--seed runs. Grouped methods each incur the listed cost. Tokens sum logged input tokens at measured per-call rates: executor 774, verifier 151, optimizer 3.4k. Shared initialization is executed once and reused by every arm.}
\label{tab:cost}
\centering\footnotesize\setlength{\tabcolsep}{3pt}
\resizebox{\linewidth}{!}{\begin{tabular}{lrrrrrr}
\toprule
\rowcolor{vchead}Method & Proposals & Gate evals & Exec. calls & Verif. calls & Optim. calls & Tokens\\
\midrule
Shared initialization & 0 & 8 & 0.7k & 0 & 0 & 0.5M\\
Text-only / Interface-only & 250 & 200 & 45.2k & 38.8k & 384 & 42.2M\\
\rowcolor{vcjoint}ViewCraft & 250 & 208 & 47.0k & 40.4k & 384 & 43.8M\\
Alternating / Staged / Indep. & 250 & 200 & 45.2k & 38.8k & 384 & 42.2M\\
MPO-adapted / MPO-fullspace & 250 & 200 & 45.2k & 38.8k & 384 & 42.2M\\
Dynamo-style / Dynamo-gen & 250 & 200 & 45.4k & 38.8k & 384 & 42.4M\\
\bottomrule\end{tabular}}
\end{table}
\paragraph{Cost accounting.}
For the ViewCraft arm, the logged call counts give
\[
47.0\mathrm{k}\times774+40.4\mathrm{k}\times151+384\times3.4\mathrm{k}
\simeq36.4\mathrm{M}+6.1\mathrm{M}+1.3\mathrm{M}=43.8\mathrm{M}.
\]
This is 1.8M input tokens per task--seed run. Arm costs differ by under 4\%; additional image-bearing gate evaluations account for the joint arm's higher cost.

\begin{table}[htbp]
\caption{Development-evaluation accounting.}
\label{tab:accounting}
\centering\footnotesize
\resizebox{0.5\linewidth}{!}{%
\begin{tabular}{l r}
\toprule
\rowcolor{vchead}Submitted (incl.\ 21 projections) & 750 \\
Pre-gate rejects (64 inert / 78 guards) & 142 \\
Paired calls (204 accepts / 404 rejects) & 608 \\
\textbf{Candidate evaluations} & \textbf{608} \\
\bottomrule
\end{tabular}}
\end{table}

\paragraph{Search-log scope and accounting.}
Each arm runs $8\times3=24$ task--seed searches. Every search has four rounds, up to four submissions per round, and 96 training rollouts. Submissions total 250 per arm, with 200--208 paired gate evaluations. The three main arms submit 750 candidates out of 1,152 available slots (65\%) and perform 608 paired evaluations. Table~\ref{tab:ledger-detail} resolves these totals by arm.

The gate accepts 204 candidates and rejects 404; accepted candidates include all 21 interface projections. Eight shared round-zero evaluations are outside these candidate counts. Separate ledgers record the candidate accounting for each strategy and ablation variant.

\begin{table}[htbp]
\caption{Submission and filtering ledger for the three main arms. Ordinary and projection columns sum to each arm's submissions or evaluations. This aggregate excludes strategy and ablation variants.}
\label{tab:ledger-detail}
\centering\footnotesize\setlength{\tabcolsep}{6pt}
\begin{tabular}{lrrrrrr}
\toprule
\rowcolor{vchead}& \multicolumn{2}{c}{Submitted} & \multicolumn{2}{c}{Pre-filtered} & \multicolumn{2}{c}{Evaluated}\\
& Ordinary & Projection & Inert & Guards & Ordinary & Projection\\
\midrule
Joint & 229 & 21 & 30 & 12 & 187 & 21\\
Text-only & 250 & 0 & 0 & 50 & 200 & 0\\
Interface-only & 250 & 0 & 34 & 16 & 200 & 0\\
\midrule
\rowcolor{vcjoint}Total & 729 & 21 & 64 & 78 & 587 & 21\\
\bottomrule
\end{tabular}
\end{table}
Projection candidates reuse validated visual edits and bypass redundant rendering replay; they still consume submission slots and a new paired evaluation. Evaluation counts therefore reflect filtering under matched submissions. All evolving methods share four rounds and 96 training rollouts. Capping joint evaluations at 200 yields mean score 0.691 versus 0.692 for full ViewCraft.

\paragraph{Reproducibility.} ReasonMap has 1{,}008 route questions over 30 official city-transit diagrams, covering fewest-transfer, no-transfer, and station-order prompts. Appendix~\ref{app:skill} typesets the exported skill. Search reproduction also requires starting skills, optimizer and verifier prompts, the expression interpreter, split manifests, and the corresponding per-item predictions.

\paragraph{Implementation.} Executor: gemini-3.8-flash, greedy decoding,
one call per observation step; budgets per domain: 12--24 steps, 4--9
images per request. Optimizer: the same model, temperature 0.7, up to 4
attempts per round, the complete executed skill in Appendix~\ref{app:skill}. Verifier for
$C_{\mathrm{ground}}$: the same model in verification mode (binary
yes/no per requirement, options withheld), two calls per gate item.
Stopping: fixed 4 rounds. All conditions share these settings. Cross-executor transfers and the case-study instance use the run with the median gate score among three seeds; both the results in the main table and those on the self-transfer diagonal pool all three runs.

\paragraph{States and comparisons.}
Let $w_0=(\theta_0,\phi_0,\psi_0)$ be Initial skill and $w_1=(\theta_1,\phi_1,\psi_1)$ the terminal joint state. Text-only independently optimizes a reader $\theta_T$, evaluating $(\theta_T,\phi_0,\psi_0)$. Interface-only independently optimizes $(\phi_V,\psi_V)$, evaluating $(\theta_0,\phi_V,\psi_V)$. The terminal $\theta$-only row evaluates $(\theta_1,\phi_0,\psi_0)$; $(\phi,\psi)$-only evaluates $(\theta_0,\phi_1,\psi_1)$. Both restore unused coordinates to the initial state. Under the clean partition, Interface-only excludes reasoning and decision rules; it covers the rendering operations, view metadata, and surface formats used by the interface.

For binary gate scores, one changed outcome moves $\Delta R$ by $0.4$--$2.1$ percentage points at the default gate sizes. Candidate and incumbent use the same questions for both score and support differences.

\paragraph{Scoring and cost.}
Table~\ref{tab:splits} distinguishes option-letter accuracy, route accuracy, field-value ANLS, ANLS, and mean LLM-Match. Four benchmarks use option letters; ReasonMap uses route matching. The input-token count for each item includes the tokens from all of its requests.

\paragraph{Semantic content and answer format.}
The contract-frozen control and format-normalized rescore in Table~\ref{tab:evidence} separate evidence use from answer presentation. Case records still expose local failures: omitting a field label changes the requested content, while a numerical value in place of an option letter can fail the answer contract despite a correct calculation.

\FloatBarrier
\section{Uncertainty and contract controls}
\label{app:uncertainty}\label{app:stats}
\begin{table}[htbp]
\caption{Joint versus alternating search. Gaps and 95\% seed-stratified media-cluster bootstrap CIs are in points. Permutation tests exchange whole source media (10,000 resamples), using per-item means over three runs.}
\label{tab:joint-alt}
\centering\footnotesize
\setlength{\tabcolsep}{5pt}
\begin{tabular}{lccr@{\hspace{16pt}}lccr}
\toprule
\rowcolor{vchead}Benchmark & Gap & 95\% CI & $p$ & Benchmark & Gap & 95\% CI & $p$\\
\midrule
MMComp & 1.3 & [0.1, 2.6] & 0.06 & VideoHolmes & 1.1 & [0.0, 2.4] & 0.07\\
ReasonMap & 1.3 & [0.1, 2.6] & 0.19 & SlideVQA & 0.9 & [$-$0.2, 2.1] & 0.12\\
FunSD & 2.1 & [0.8, 3.5] & 0.01 & OpenEQA & 1.2 & [0.2, 2.5] & 0.09\\
STI-Bench & 1.8 & [0.5, 3.3] & 0.04 & MMIU & 1.1 & [0.0, 2.3] & 0.08\\
\midrule
\rowcolor{vcjoint}\multicolumn{8}{c}{Mean gap: 1.3 points\quad 95\% CI: [0.6, 2.1]}\\
\bottomrule
\end{tabular}
\end{table}
\paragraph{Baseline and projection comparisons.}
Joint search leads alternating search in every seed, with gaps of $+1.5$, $+1.2$, and $+1.3$ points. Table~\ref{tab:joint-alt} reports the task-level comparison.

ViewCraft's gain over MPO-fullspace has CI $[+1.3,+3.9]$; all tasks favor ViewCraft (sign-test $p=0.008$). The same-space margins over Dynamo-gen and random proposals appear in Table~\ref{tab:ladder}.

\begin{table}[htbp]
\caption{Mean task-score increments along the matched-budget ladder, and final margins. All differences are in points.}
\label{tab:ladder}\centering\footnotesize
\begin{tabular}{lr@{\hspace{18pt}}lr}\toprule
\rowcolor{vchead}Ladder step & Gain & ViewCraft comparison & Margin\\\midrule
Prompt space to random same-space & +0.7 & MPO-fullspace & +2.6\\
Random to UCB selection & +0.5 & Dynamo-gen & +3.0\\
UCB to failure-guided joint & +1.0 & Random proposals & +3.1\\
Gate-only to full retention & +1.6 & Random to failure-guided joint & +1.5\\
\bottomrule\end{tabular}\end{table}

\begin{table}[htbp]
\caption{Uncertainty versus Direct. Std is over three search runs. Gain CIs are 95\% seed-stratified media-cluster bootstraps; permutation tests use each item's mean score over the three runs. Gains and intervals are in points.}
\label{tab:uncertainty}
\centering\footnotesize
\resizebox{0.8\linewidth}{!}{%
\begin{tabular}{l c c c c}
\toprule
\rowcolor{vchead}Benchmark & ViewCraft (std) & Gain (pts) & Cluster CI (pts) & vs.\ Direct $p$ \\
\midrule
MMComp & 0.874$\pm$0.007 & 4.0 & [0.3, 7.8] & 0.03 \\
ReasonMap & 0.898$\pm$0.008 & 4.0 & [0.2, 7.9] & 0.05 \\
FunSD & 0.412$\pm$0.015 & 10.6 & [5.8, 15.2] & $<$0.001 \\
STI-Bench & 0.496$\pm$0.011 & 8.6 & [2.9, 14.2] & 0.006 \\
VideoHolmes & 0.472$\pm$0.010 & 6.4 & [1.2, 11.6] & 0.01 \\
SlideVQA & 0.858$\pm$0.007 & 5.6 & [0.1, 11.1] & 0.05 \\
OpenEQA & 0.574$\pm$0.014 & 6.2 & [1.6, 10.9] & 0.02 \\
MMIU & 0.951$\pm$0.005 & 3.0 & [0.0, 6.0] & 0.08 \\
\bottomrule
\end{tabular}}

\end{table}
For MMIU, Direct's score of 0.921 leaves 7.9 points of headroom; ViewCraft recovers 3.0 points.

\paragraph{Significance testing.}
All tests operate on per-item means over the three runs. Paired permutation tests exchange whole source media over 10{,}000 resamples of these per-item means. Confidence intervals use seed-stratified media-cluster bootstrap resampling: whole films, forms, or decks are resampled within seed strata and then averaged. Search-run standard deviations are reported separately. Binary and continuous metrics are evaluated with this same per-item statistical protocol.

All task-level estimates favor ViewCraft (exact sign test $p=0.008$). Across-task mean and median gains are $+6.1$ and $+5.9$ points. After applying Holm correction across tasks, the individually significant gains are those on FunSD and STI-Bench (Table~\ref{tab:gain-summary}).

\begin{table}[htbp]
\caption{Across-task gain summaries and multiplicity correction versus Direct.}
\label{tab:gain-summary}\centering\footnotesize
\begin{tabular}{lr@{\hspace{18pt}}lr}\toprule
\rowcolor{vchead}Summary & Value & Holm-adjusted comparison & $p$\\\midrule
Mean gain (points) & +6.1 & FunSD & 0.004\\
Median gain (points) & +5.9 & STI-Bench & 0.042\\
Relative gain range (\%) & 3.3--34.6 & Other tasks & $>0.05$\\
Median relative gain (\%) & 9.5 & & \\
\bottomrule\end{tabular}\end{table}

\begin{table}[htbp]
\caption{Projection design at matched budget (per arm: 250 submitted
proposals). Gap is full ViewCraft minus the variant in points; CIs are
95\% seed-stratified media-cluster bootstraps of the paired per-item
difference; $p$ is a paired permutation test on per-item means (10{,}000
resamples).}
\label{tab:projdesign}
\centering\footnotesize
\setlength{\tabcolsep}{3.2pt}
\resizebox{\linewidth}{!}{%
\begin{tabular}{l cccccccc c c c}
\toprule
\rowcolor{vchead}
Variant & MM & RMap & FunSD & STI & VH & Slide & OEQA & MMIU & Mean & Gap [CI] & $p$\\
\midrule
No retry (drop on joint rejection) & 0.870 & 0.893 & 0.407 & 0.483 & 0.461 & 0.849 & 0.565 & 0.944 & 0.684 & 0.8 [0.3, 1.3] & 0.004\\
Plain retry (fresh proposals) & 0.872 & 0.893 & 0.408 & 0.484 & 0.463 & 0.850 & 0.566 & 0.944 & 0.685 & 0.7 [0.2, 1.2] & 0.008\\
Symmetric projection (keep $\theta'$) & 0.872 & 0.895 & 0.409 & 0.486 & 0.463 & 0.851 & 0.567 & 0.946 & 0.686 & 0.6 [0.1, 1.1] & 0.020\\
No $\Delta C$ trigger (retry all eligible) & 0.872 & 0.895 & 0.408 & 0.487 & 0.464 & 0.852 & 0.568 & 0.947 & 0.687 & 0.5 [0.1, 1.0] & 0.030\\
\textbf{Asymmetric projection (ours)} & \textbf{0.874} & \textbf{0.898} & \textbf{0.412} & \textbf{0.496} & \textbf{0.472} & \textbf{0.858} & \textbf{0.574} & \textbf{0.951} & \textbf{0.692} & --- & ---\\
\bottomrule
\end{tabular}}
\end{table}

\begin{table}[htbp]
\caption{Isolating evidence from contract effects. ``Evidence-only''
freezes the answer and protocol contracts at their initial values and
evolves only evidence selection, transformation, layout, and the paired
reading rules. ``Format-normalized'' rescores all responses
case-/unit-insensitively with letter extraction.}
\label{tab:evidence}
\centering\footnotesize
\setlength{\tabcolsep}{3pt}
\resizebox{\linewidth}{!}{%
\begin{tabular}{l cccccccc}
\toprule
\rowcolor{vchead}Condition & MMComp & RMap & FunSD & STI & VHolm & Slide & OEQA & MMIU \\
\midrule
ViewCraft & 0.874 & 0.898 & 0.412 & 0.496 & 0.472 & 0.858 & 0.574 & 0.951 \\
\quad format-normalized & 0.876 & 0.900 & 0.421 & 0.498 & 0.474 & 0.861 & 0.581 & 0.953 \\
Evidence-only (contracts frozen) & 0.872 & 0.895 & 0.401 & 0.492 & 0.468 & 0.854 & 0.566 & 0.947 \\
\quad format-normalized & 0.872 & 0.895 & 0.409 & 0.495 & 0.470 & 0.857 & 0.574 & 0.947 \\
\bottomrule
\end{tabular}}

\end{table}
The contract-frozen condition preserves most of ViewCraft's gain. Format-normalized rescoring keeps visual inputs and generated responses fixed while applying a common scoring rule. On FunSD, Direct gains 1.2 points and ViewCraft 0.9 points, retaining ViewCraft's advantage.

\Needspace{2\baselineskip}
Figure~\ref{fig:cases-boundaries} distinguishes preservation of the content requested by the question from compliance with all formatting requirements of the complete response.

\begin{table}[htbp]
\caption{Fixed-trajectory decomposition. Each row replays the terminal
skill's recorded observation sequence and substitutes one coordinate. The
evidence transplant delivers the evolved views to the initial reader; the
reading transplant gives the evolved reader the initial views.}
\label{tab:fixedtraj}
\centering\footnotesize\setlength{\tabcolsep}{3pt}
\resizebox{\linewidth}{!}{%
\begin{tabular}{l cccccccc c}
\toprule
\rowcolor{vchead}
Replayed state & MMComp & RMap & FunSD & STI & VHolm & Slide & OEQA & MMIU & Mean \\
\midrule
Initial skill $(\theta_0,v_0)$ & 0.786 & 0.841 & 0.288 & 0.385 & 0.381 & 0.812 & 0.478 & 0.908 & 0.610 \\
Evidence transplant $(\theta_0,v_1)$ & 0.775 & 0.845 & 0.318 & 0.440 & 0.424 & 0.823 & 0.519 & 0.922 & 0.633 \\
Reading transplant $(\theta_1,v_0)$ & 0.848 & 0.861 & 0.345 & 0.410 & 0.403 & 0.829 & 0.503 & 0.918 & 0.640 \\
\rowcolor{vcjoint}\textbf{ViewCraft $(\theta_1,v_1)$} & \textbf{0.874} & \textbf{0.898} & \textbf{0.412} & \textbf{0.496} & \textbf{0.472} & \textbf{0.858} & \textbf{0.574} & \textbf{0.951} & \textbf{0.692} \\
\bottomrule
\end{tabular}}
\end{table}
With the observation trajectory fixed, the evidence and reading transplants each recover part of the joint gain. Table~\ref{tab:fixed-gaps} reports their additive contrast; the terminal crossing in Figure~\ref{fig:joint-necessity}c allows each reader to issue its own requests.
\begin{table}[htbp]\caption{Fixed-trajectory contrasts relative to Initial skill, in points. The final row is the joint gain minus the two separate transplant gains.}\label{tab:fixed-gaps}
\centering\footnotesize\begin{tabular}{lr}\toprule
Evidence transplant & +2.3\\Reading transplant & +3.0\\Joint pairing & +8.2\\Joint-minus-transplants contrast & +2.9\\\bottomrule
\end{tabular}\end{table}
\begin{table}[htbp]
\caption{Terminal interaction $I$ (points) under three coordinate
partitions. Storage keeps exported files intact; clause-content assigns
inference and decision rules to $\theta$; strict additionally moves
render-referencing clauses (note placement, rendered-color references)
to the interface.}
\label{tab:partition}
\centering\footnotesize\setlength{\tabcolsep}{3pt}
\begin{tabular}{l cccccccc c}
\toprule
\rowcolor{vchead}Partition & MM & RMap & FunSD & STI & VH & Slide & OEQA & MMIU & Mean\\
\midrule
Storage fields & 1.4 & 1.0 & 1.6 & 0.8 & 0.6 & $-$0.3 & 0.5 & 0.8 & +0.8\\
Strict (render clauses out of $\theta$) & 1.8 & 1.5 & 1.8 & 1.2 & 1.0 & 0.5 & 1.2 & 0.9 & +1.2\\
\rowcolor{vcjoint}Clause-content (used throughout) & 2.4 & 2.2 & 2.3 & 1.7 & 1.5 & 0.9 & 1.8 & 1.2 & +1.8\\
\bottomrule
\end{tabular}
\end{table}
\begin{table}[htbp]
\caption{Gate-split sensitivity. Differences use the displayed mean task scores; the final row reports joint minus alternating in points.}
\label{tab:altsplit}\centering\footnotesize
\begin{tabular}{lrrr}\toprule
\rowcolor{vchead}Strategy & Split A & Split B & Difference (points)\\\midrule
Alternating & 0.678 & 0.674 & $-$0.4\\
Gate-only joint & 0.676 & 0.672 & $-$0.4\\
ViewCraft & 0.692 & 0.688 & $-$0.4\\\midrule
Joint $-$ Alternating (points) & +1.4 & +1.4 & 0.0\\\bottomrule
\end{tabular}\end{table}
Re-drawing the gate items shifts every arm by under half a point and preserves the joint-over-alternating gap computed from the displayed means (Table~\ref{tab:altsplit}).

\FloatBarrier
\section{Transfer and search diagnostics}
\label{app:search-details}
\begin{figure}[htbp]
\centering
\includegraphics[width=0.70\linewidth]{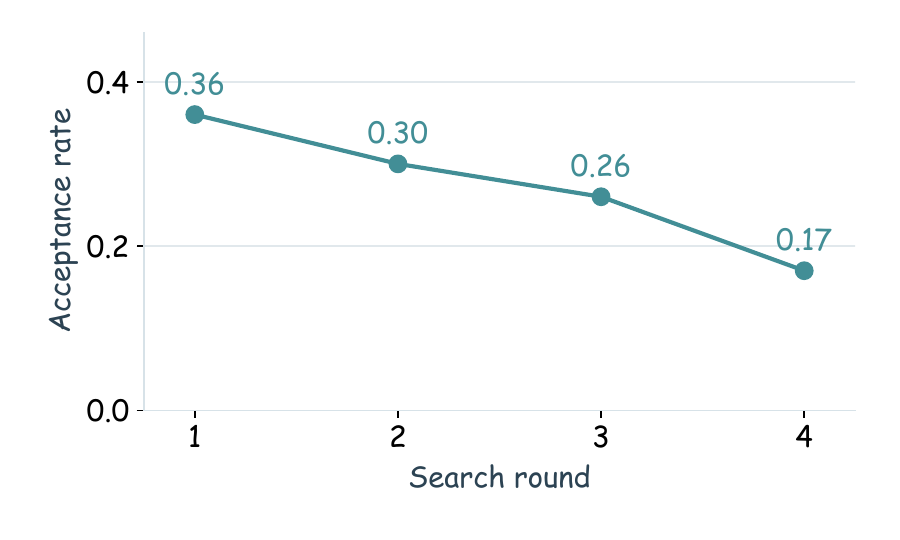}
\caption{\textbf{Search selectivity.} Rates pool the three evolving arms and use submitted proposals, including projection retries and pre-gate rejects. Rejected replays and guard failures consume no development calls.}
\label{fig:search}
\end{figure}

Every evaluated transferred skill improves over its target executor's Direct baseline by at least 1.5 points. Three transfers also exceed the target-evolved skill; the remaining transfers fall within 3.3 points of it. The tightest margin over Direct is GPT-5.5 executing MMComp with the Sonnet skill.

Acceptance declines across search rounds, while score-drop rejections identify candidates that underperform the incumbent (Figure~\ref{fig:search}). These diagnostics characterize selection during search; the held-out comparisons establish the value of the retained skills. The most frequently accepted co-edits pair revisions to the reader with revisions to the view contract (Figure~\ref{fig:coedits}a).

Within the recorded accepted cohort (including projection), 153 of 204 edits both repair and regress items. The averages are 5.4 repairs and 2.9 regressions. In this recorded cohort, the paired gate selects a net improvement even when some previously solved items regress.

The rejected FunSD proposal returned ``None'' on 89\% of paired items, reducing the score from 0.273 to 0.075. Its demand for a literal query-label match excluded valid composite fields. The next proposal relaxed this literal-match requirement through fuzzy matching.

\FloatBarrier
\section{Discrete proposal space}
\label{app:details}
The clean partition assigns all inference and decision rules to $\theta$: polarity handling, verdict-table logic, when-to-zoom decisions, and mark-reading procedures. The interface contains executable rendering operations in $\phi$ (selection, transformation, crop admission, canvas), layout and view metadata in $\psi$, and answer/action surface formats. Crop admission controls whether a requested crop is rendered; deciding whether to request it belongs to the reader. The terminal crossing restores all reasoning clauses with $\theta$, including those stored in interface files.

Clause-content assigns inference and decision rules to $\theta$; mark placement and names of rendered colors belong to the interface vocabulary. Storage keeps exported files intact. The strict partition moves every render-referencing clause to the interface before the crossing is computed (Table~\ref{tab:partition}).

In the exported instance, \texttt{answer\_contract} contains both the single-letter format and decision rules for choosing that letter; \texttt{protocol\_contract} contains both JSON action syntax and domain guidance. The former's polarity/verdict clauses and the latter's reading/zoom policy belong to $\theta$ under the re-partition. View descriptions, color identities, and action schemas remain in the interface. 

\FloatBarrier
\section{Reading the recorded cases}
\begin{figure}[htbp]
\centering\includegraphics[width=\linewidth]{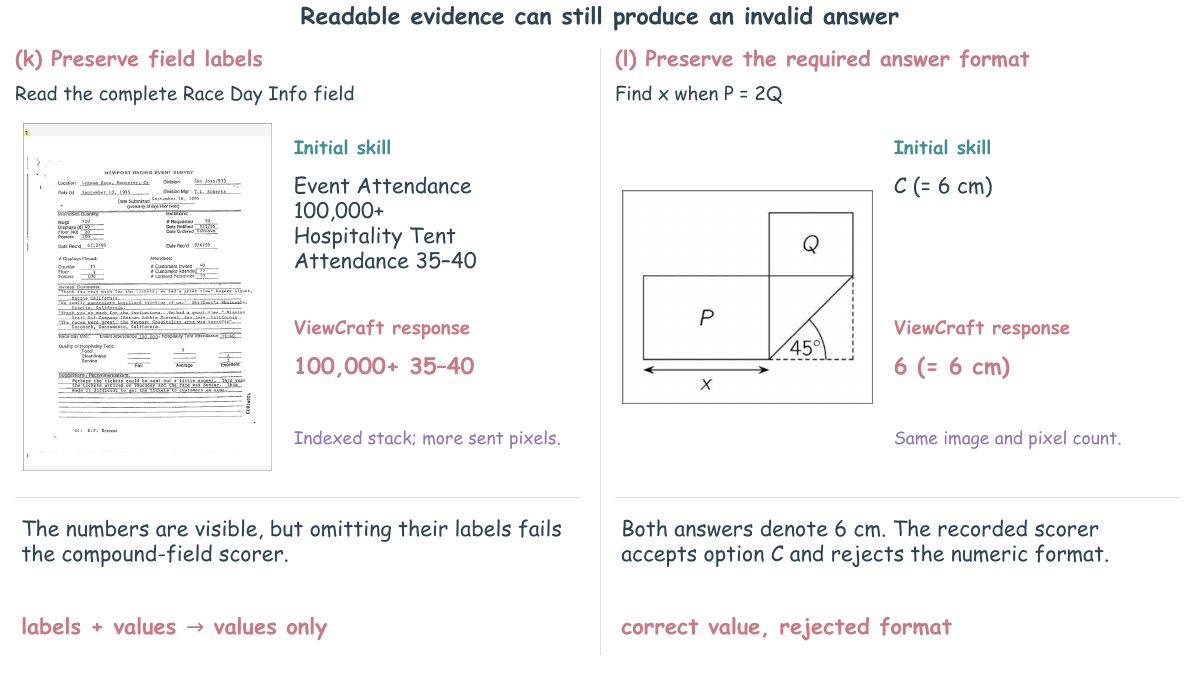}
\caption{\textbf{Two answer-contract regressions.} In (k), the response retains the numbers but drops the labels required by the compound-field query. In (l), both responses denote 6 cm, yet only option letter C passes the recorded scorer.}
\label{fig:cases-boundaries}
\end{figure}

Figure~\ref{fig:cases-overview} presents ten repairs and Figure~\ref{fig:cases-boundaries} two regressions. The figures pair questions and responses with the corresponding view changes and reading rules behind them.

Long video stacks are reflowed or excerpted for legibility, with their source timestamps retained. Each displayed view comes from evidence sent to the executor.

\FloatBarrier
\section{Preserving query semantics and answer format}
In (k), the evolved response keeps both numbers but drops field labels required by the compound-field scorer. In (l), both conditions read 6\,cm, but only a response using the option letter passes the recorded scorer. These cases distinguish preserving the requested field associations from producing an admissible answer format, motivating evaluation of the complete response.

\Needspace{10\baselineskip}
\FloatBarrier
\section{Absolute input-token costs}
\begin{figure}[htbp]
\centering
\includegraphics[width=\linewidth]{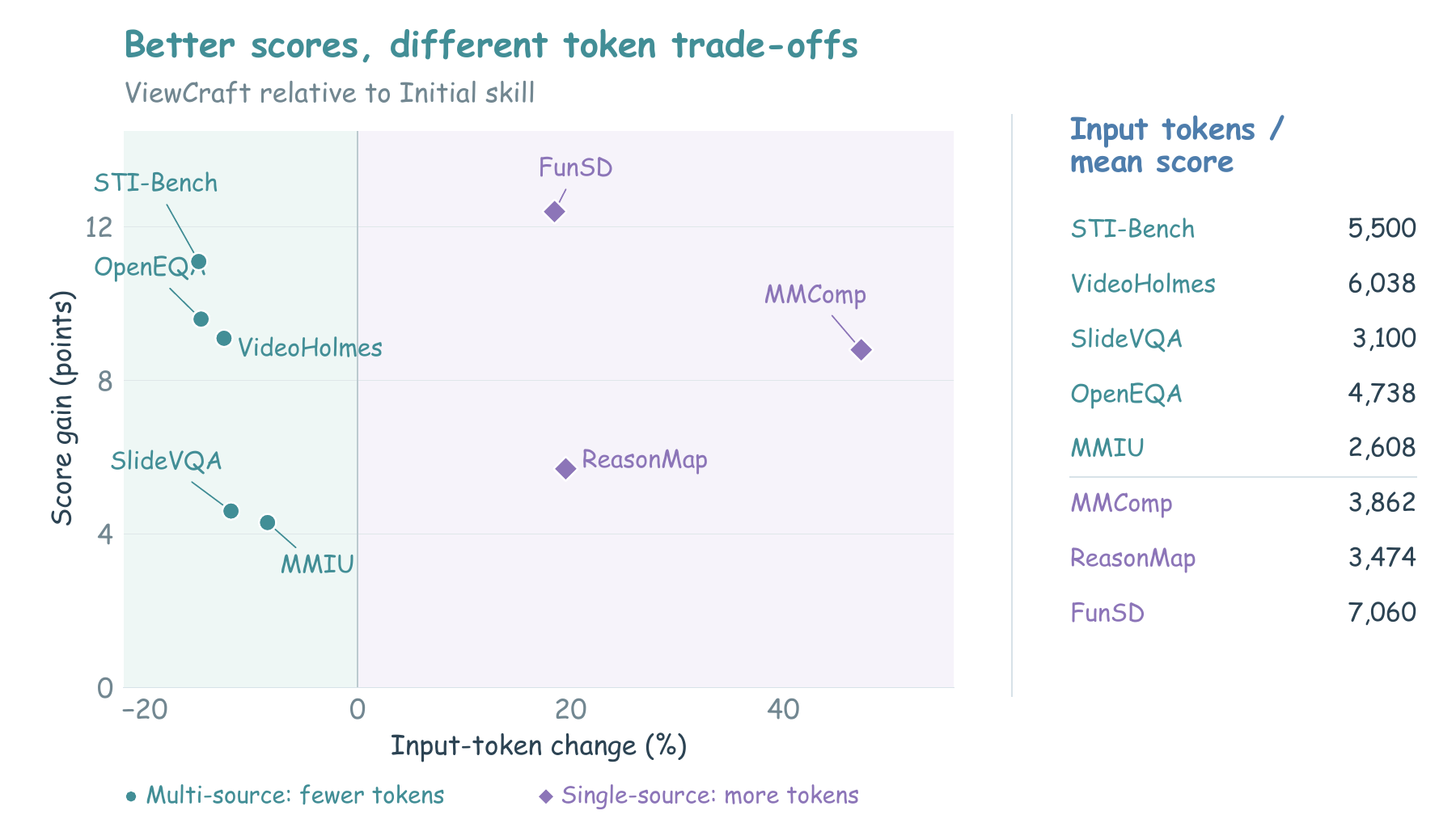}
\caption{\textbf{Task scores and input-token cost.} Left: score gains and token changes relative to Initial skill; negative horizontal values mean savings. Circles denote multi-source tasks, diamonds single-source tasks. Right: ViewCraft input tokens divided by mean score (a score-normalized cost for continuous metrics). Counts include all requests; Appendix Table~\ref{tab:token-counts} lists absolute counts.}
\label{fig:tokens}\label{fig:tradeoff}
\end{figure}

Figure~\ref{fig:tokens} compares task scores and input-token costs.
\begin{table}[htbp]
\caption{Mean input tokens per item across all requests. Original counts determine Figure~\ref{fig:tokens} and the appended ViewCraft-minus-Direct column. Relative to Direct, increases are $15$--$23\%$ for multi-source tasks and $67$--$108\%$ for single-source tasks; multi-source counts fall approximately $8$--$15\%$ from Initial skill.}
\label{tab:token-counts}
\centering\small\setlength{\tabcolsep}{8pt}\renewcommand{\arraystretch}{1.1}
\begin{tabular}{lrrrr}\toprule
Benchmark & Direct & Initial skill & ViewCraft & VC $-$ Direct\\\midrule
STI-Bench & 2,380 & 3,210 & 2,730 & +350\\
VideoHolmes & 2,410 & 3,260 & 2,850 & +440\\
SlideVQA & 2,240 & 3,020 & 2,660 & +420\\
OpenEQA & 2,360 & 3,190 & 2,720 & +360\\
MMIU & 2,010 & 2,710 & 2,480 & +470\\
MMComp & 1,620 & 2,291 & 3,375 & +1,755\\
ReasonMap & 1,850 & 2,610 & 3,120 & +1,270\\
FunSD & 1,740 & 2,456 & 2,910 & +1,170\\
\bottomrule\end{tabular}\end{table}
\FloatBarrier
\begin{table}[htbp]
\caption{Deployment cost averaged equally over the eight tasks. Input means are the arithmetic means of Table~\ref{tab:token-counts}; search cost is amortized over reuse.}
\label{tab:deploy}
\centering\footnotesize
\setlength{\tabcolsep}{3pt}
\resizebox{0.8\linewidth}{!}{%
\begin{tabular}{l c c c c}
\toprule
\rowcolor{vchead}
& Input tok/item & Output tok/item & Calls/item & Latency (s/item) \\
\midrule
Direct & 2.08k & 38 & 1.0 & 4.2 \\
Initial skill & 2.84k & 61 & 2.6 & 9.4 \\
\rowcolor{vcjoint}ViewCraft & 2.86k & 69 & 3.4 & 11.8 \\
\bottomrule
\end{tabular}}
\end{table}
Reuse amortizes the one-time search cost in Table~\ref{tab:cost}. On multi-source tasks, input selection reduces input tokens relative to Initial skill. ViewCraft adds 7.6 seconds per item relative to Direct and generates more output tokens than Initial skill. Table~\ref{tab:deploy} reports the arithmetic mean of the eight per-task input counts in Table~\ref{tab:token-counts}, alongside output tokens, calls, and end-to-end latency.

\begin{table}[htbp]
\caption{Task scores at matched and expanded input budgets. ``Token-matched'' gives Direct
ViewCraft's measured per-item input budget through content-independent
means (more frames or pages on multi-source tasks; upsampling plus a
regular crop grid on single-source tasks). ``$\times$3'' is
self-consistency over three samples. Tokens/item is the task-equal-weight
macro-average.}
\label{tab:budget}
\centering\footnotesize\setlength{\tabcolsep}{3pt}
\resizebox{\linewidth}{!}{%
\begin{tabular}{l cccccccc c c}
\toprule
\rowcolor{vchead}
& MMComp & RMap & FunSD & STI & VHolm & Slide & OEQA & MMIU & Mean & Tok/item \\
\midrule
Direct & 0.834 & 0.858 & 0.306 & 0.410 & 0.408 & 0.802 & 0.512 & 0.921 & 0.631 & 2.08k \\
Direct (token-matched) & 0.841 & 0.863 & 0.318 & 0.421 & 0.417 & 0.809 & 0.521 & 0.926 & 0.640 & 2.86k \\
Direct $\times$3 (self-consistency) & 0.848 & 0.867 & 0.311 & 0.418 & 0.413 & 0.808 & 0.519 & 0.929 & 0.639 & 6.24k \\
Initial skill & 0.786 & 0.841 & 0.288 & 0.385 & 0.381 & 0.812 & 0.478 & 0.908 & 0.610 & 2.84k \\
\rowcolor{vcjoint}\textbf{ViewCraft} & \textbf{0.874} & \textbf{0.898} & \textbf{0.412} & \textbf{0.496} & \textbf{0.472} & \textbf{0.858} & \textbf{0.574} & \textbf{0.951} & \textbf{0.692} & 2.86k \\
\bottomrule
\end{tabular}}
\end{table}
At identical input-token budget, ViewCraft leads Direct by $+5.2$ points, with CI $[+3.6,+6.9]$ and positive gains on every task. Aggregating three samples from Direct through self-consistency also yields a similar mean task score; Table~\ref{tab:budget} reports the corresponding comparison.

\FloatBarrier
\section{Gate definitions and round procedure}
\label{app:mechanics}
\begin{table}[htbp]
\caption{Mechanism ablations across eight tasks and three seeds at 250 submissions per arm. Paired intervals and tests follow Table~\ref{tab:projdesign}. Gate evaluations: full 208 (21 projections; 30 inert and 12 guard failures filtered), no replay 238, no projection 204, no tiebreak 208, and gate-only 200.}
\label{tab:ablations}
\centering\footnotesize
\setlength{\tabcolsep}{3pt}
\resizebox{\linewidth}{!}{%
\begin{tabular}{l cccccccc c c c}
\toprule
\rowcolor{vchead}Variant & MM & RMap & FunSD & STI & VH & Slide & OEQA & MMIU & Mean & Gap [CI] & $p$\\
\midrule
w/o replay filter & 0.870 & 0.891 & 0.404 & 0.487 & 0.463 & 0.848 & 0.562 & 0.942 & 0.683 & 0.9 [0.4, 1.5] & 0.002\\
w/o interface projection & 0.870 & 0.893 & 0.407 & 0.483 & 0.461 & 0.849 & 0.565 & 0.944 & 0.684 & 0.8 [0.3, 1.3] & 0.004\\
w/o evidence tiebreak & 0.870 & 0.891 & 0.403 & 0.489 & 0.464 & 0.852 & 0.566 & 0.947 & 0.685 & 0.7 [0.2, 1.2] & 0.008\\
Gate-only joint (all three off) & 0.861 & 0.887 & 0.393 & 0.478 & 0.457 & 0.844 & 0.560 & 0.932 & 0.676 & 1.6 [0.8, 2.4] & $<$0.001\\
\rowcolor{vcjoint}\textbf{ViewCraft (full)} & \textbf{0.874} & \textbf{0.898} & \textbf{0.412} & \textbf{0.496} & \textbf{0.472} & \textbf{0.858} & \textbf{0.574} & \textbf{0.951} & \textbf{0.692} & --- & ---\\
\bottomrule
\end{tabular}}
\end{table}
The optimizer extracts at most three requirements from the question alone. For item $i$, let $H_i$ contain these atomic requirements and let $s_{ij}\in\{0,1\}$ be the verifier's support judgment for requirement $j$ given the actual sent views. For a nonempty requirement set,
\begin{equation}
C_i(w)=\frac{1}{|H_i|}\sum_{j\in H_i}s_{ij}(w),\qquad
\Delta C=\frac{1}{|G|}\sum_{i\in G}[C_i(c)-C_i(w)].
\end{equation}
The same requirements and gate items are used for both states; reference answers are withheld from extraction and verification. Two annotators independently audited 100 stratified judgments, with pre-consensus Cohen's $\kappa=0.76$, then resolved disagreements by discussion. The frozen verifier agreed with the resulting consensus on 89\% of judgments. The tiebreak's 44 accepts concentrate on visual edits that pass replay. Removing it lowers task scores (Table~\ref{tab:ablations}); this ablation measures the rule's effect, while the annotation audit measures verifier agreement.

Replay checks rendered changes for proposals that edit the visual side. Reader-only and answer-text-only proposals bypass the pixel filter. This scope permits the solo reader accepts shown in Figure~\ref{fig:coedits}a. Execution checks still apply to every candidate before scoring.

Projection restores the incumbent reader while retaining a rejected candidate's visual edits. It is eligible after a task-score loss with $\Delta C\ge0$ and requires both submission and gate allowances. The 21 accepted retries are included in the 204 accepted edits and in the 750 submitted candidates. Algorithm~\ref{alg:round} takes the remaining gate and submission allowances as inputs.

\begin{algorithm}[htbp]
\caption{One ViewCraft round with at most four submissions, including projection retries. Each paired evaluation consumes one gate unit; pre-evaluation rejections consume no gate units (Table~\ref{tab:accounting}).}
\label{alg:round}
\begingroup
\small
\setstretch{1.0}
\begin{algorithmic}[1]
\Require Incumbent $w$ with cached gate scores, training traces $T$, gate items $G$, remaining gate allowance $B$, remaining submissions $P$, strategy $S$
\State $L\gets 4$ \Comment{Remaining submission slots in this round}
\While{$L>0$ and $P>0$ and $B>0$}
  \State $b\gets w$; propose $c$ from $b,T,S$; $L\gets L-1$; $P\gets P-1$
  \If{$c$ changes rendering fields and replay is pixel-identical}
    \State Record $(b,c,\mathrm{rejected\mbox{-}inert})$ in $S$; \textbf{continue}
  \EndIf
  \If{$c$ fails $C_{\mathrm{exec}}$}
    \State Record $(b,c,\mathrm{rejected\mbox{-}guard})$ in $S$; \textbf{continue}
  \EndIf
  \State Evaluate $c$ against $b$ on $G$; obtain $\Delta R,\Delta C$; $B\gets B-1$
  \If{$\Delta R>0$ or $(\Delta R=0$ and $\Delta C>0)$}
    \State Record $(b,c,\mathrm{accepted})$ in $S$; $w\gets c$; refresh cached scores
    \State \textbf{break}
  \EndIf
  \State Record $(b,c,\mathrm{rejected\mbox{-}score/support})$ in $S$
  \If{$c$ has joint reading/visual edits, $\Delta R<0$, $\Delta C\ge0$, and $L>0$, $P>0$, $B>0$}
    \State $c_v\gets(\theta_b,\phi_c,\psi_c)$; $L\gets L-1$; $P\gets P-1$; reuse validated rendering
    \If{$c_v$ fails $C_{\mathrm{exec}}$}
      \State Record $(b,c_v,\mathrm{rejected\mbox{-}guard})$ in $S$; \textbf{continue}
    \EndIf
    \State Evaluate $c_v$ against $b$ on $G$; obtain $\Delta R_v,\Delta C_v$; $B\gets B-1$
    \If{$\Delta R_v>0$ or $(\Delta R_v=0$ and $\Delta C_v>0)$}
      \State Record $(b,c_v,\mathrm{accepted})$ in $S$; $w\gets c_v$; refresh cached scores
      \State \textbf{break}
    \Else
      \State Record $(b,c_v,\mathrm{rejected\mbox{-}score/support})$ in $S$
    \EndIf
  \EndIf
\EndWhile
\State \Return $(w,B,P,S)$
\end{algorithmic}
\endgroup
\end{algorithm}

\FloatBarrier
\section{A complete evolved skill}
\label{app:skill}
The MMComp instance consists of \texttt{SKILL.md}, \texttt{VIEW.yaml}, and \texttt{LAYOUT.yaml}. These are storage containers, not the logical coordinate split: reasoning clauses in any file belong to $\theta$, as specified in Appendix~\ref{app:details}. The reader first grounds option nouns and builds a verdict table. The crop rule is \texttt{zoom\_gain > 1.05 or region\_frac < 0.12}; canvas expressions color notes by option ordinal. The view contract explains those colors, and the answer contract requires the final option letter. These files make the relation between the reader, visual configuration, and interface concrete. The complete contents are typeset below, including four runtime slots: \texttt{\{pool\_size\}}, \texttt{\{max\_crops\}}, \texttt{\{ans\_header\}}, and \texttt{\{ans\_example\}}. The domain configuration binds these at load time. For MMComp, \texttt{pool\_size}=8 source units per item and \texttt{max\_crops}=2, matching \texttt{max\_crops\_per\_step}; the protocol template supplies the answer header and example. The frozen artifact includes the template and these bindings. The Markdown and YAML files are typeset below to make them readable while preserving their original content.

\begingroup
\small
\renewcommand{\ttdefault}{zi4}
\setlength{\parindent}{0pt}
\setlength{\parskip}{2pt plus 1pt}
\setlist[itemize]{leftmargin=1.4em,itemsep=1pt,parsep=0pt,topsep=2pt}
\setlist[enumerate]{leftmargin=1.6em,itemsep=1pt,parsep=0pt,topsep=2pt}
\newcommand{\skillheading}[1]{\par\addvspace{7pt}\Needspace{4\baselineskip}\noindent{\color{vcblue}\bfseries #1}\par\nopagebreak}
\newcommand{\skilltitle}[1]{\par\noindent{\bfseries #1}\par\nopagebreak}
\tcbset{skillpanel/.style={enhanced,breakable,colback=vchead,colframe=vcblue!35,
  boxrule=0.3pt,arc=2pt,left=7pt,right=7pt,top=5pt,bottom=5pt,
  before skip=6pt,after skip=6pt}}
\Needspace{9\baselineskip}
\subsection{Reading procedure: SKILL.md}
\skilltitle{mmcomposition Reading Procedure}
\skillheading{What this domain contains}
Items are single composite images --- multi-panel layouts, presentation slides, side-by-side photo pairs, captioned images, or scene photos. The question probes spatial relationships, outlier detection, caption matching, or scene description. Answer options (A/B/C/D or Left/Right style) are always embedded in the question text you receive --- never refuse on the grounds that options are missing.

\skillheading{Step 1 --- Noun-grounding pre-pass (ALWAYS, before any other reasoning)}
Before interpreting the image content against the question, do this explicitly:

\begin{enumerate}
\item Parse each answer option and extract the key concrete nouns (objects, people, animals, scene elements).

\item For each option, state whether its key noun(s) are \textbf{visible} or \textbf{absent} in the image.

\item Count how many options describe objects that are entirely absent from the image.

\item Note the question's polarity:

\begin{itemize}
\item 'which does NOT apply' / 'pick the incorrect' / 'does not apply' / 'wrong description' \ensuremath{\rightarrow} NEGATION-FRAMED: proceed to Step 1-V (Verdict Table) immediately.

\item 'which aligns / which is correct / right description / accurately describes' \ensuremath{\rightarrow} AFFIRMATIVE: proceed to Step 1-V as well, marking each option APPLIES or ABSENT.

\item 'yes/no' \ensuremath{\rightarrow} after confirming subject locations with notes, fill in: 'The [subject] IS / IS NOT [relation] the [object]' before mapping to A/B.

\end{itemize}
\item Write this grounding check in your note field: 'Objects in options: [list]. Visible: [list]. Absent: [list]. Question asks for: [correct/incorrect/yes/no].'

\end{enumerate}
This pre-pass prevents over-elimination (answering all letters) and polarity flips.

\skillheading{Step 1-V --- MANDATORY VERDICT TABLE (for all multi-option questions)}
Before committing to any letter, write a verdict table in your note field. Format exactly:

\begin{tcolorbox}[skillpanel]
\begin{Verbatim}[fontfamily=zi4,fontsize=\footnotesize,breaklines=true,breakanywhere=true,breaksymbolleft={},breaksymbolright={},breakindent=0pt,numbers=none]
VERDICT TABLE
A: [one-line visual evidence or NOT FOUND IN IMAGE] → APPLIES / ABSENT
B: [one-line visual evidence or NOT FOUND IN IMAGE] → APPLIES / ABSENT
C: [one-line visual evidence or NOT FOUND IN IMAGE] → APPLIES / ABSENT
D: [one-line visual evidence or NOT FOUND IN IMAGE] → APPLIES / ABSENT
QUESTION POLARITY: [asks for INCORRECT / asks for CORRECT]
FINAL LETTER: [the letter whose verdict matches the polarity]
\end{Verbatim}
\end{tcolorbox}
Rules:

\begin{itemize}
\item For negation-framed questions ('does NOT apply', 'pick the incorrect', 'wrong description'): the FINAL LETTER is the one marked ABSENT (or the most specifically wrong if multiple are absent).

\item For affirmative questions ('which is correct', 'accurately describes', 'right description'): the FINAL LETTER is the one marked APPLIES.

\item If ALL options are absent (full domain mismatch): rank by specificity of false claim --- the option making the most concrete/specific false assertion is 'most incorrect'; that letter is FINAL LETTER.

\item The answer field must contain exactly the FINAL LETTER from this table. No other letter may appear in the answer field.

\item Drop one note on the image for each option (at the region where the object was sought --- even if absent, mark the empty region). Note ordinal 1=A, 2=B, 3=C, 4=D. The canvas will color each note ring distinctly so you can visually confirm letter-to-region mapping.

\end{itemize}
\skillheading{Step 2 --- Initial full read}
Attach and read image 0. Identify:

\begin{itemize}
\item How many panels/sub-images are present

\item Question type: spatial-relation \textbar{} outlier \textbar{} caption-match \textbar{} scene-description \textbar{} left-right \textbar{} mismatch-detection

\item Whether text in the image is legible at full scale

\item Whether any embedded screen/monitor/framed photo contains the relevant content (screen-within-screen pattern)

\end{itemize}
\skillheading{Step 3 --- Decide if one pass is enough}
Answer immediately (after the verdict table) when:

\begin{itemize}
\item The scene content is unambiguous at full scale AND the noun-grounding pre-pass resolves the question

\item The answer options are purely verbal and the image content is already understood

\end{itemize}
Do NOT answer immediately when:

\begin{itemize}
\item Slide or panel text is small and needs reading

\item Two panels look similar and fine detail distinguishes them

\item A spatial relationship depends on object overlap or occlusion

\item A TV/monitor/framed image is embedded and may contain the described content

\end{itemize}
\skillheading{Step 4 --- Zoom protocol (when needed)}
For \textbf{outlier / slide questions}: zoom each panel individually to read titles, headers, colour schemes.

For \textbf{caption-match}: zoom embedded text and the dominant subject.

For \textbf{left-right / spatial}: zoom the region of interest. Place a note at the key anchor before answering.

For \textbf{screen-within-screen}: detect bright rectangular inset (TV, monitor, framed photo). Zoom to that inset at the tightest crop that captures the full screen content. Treat the inset as a separate visual context and re-run the noun-grounding pre-pass on it.

For \textbf{mismatch / all-absent scenarios}: zoom the most ambiguous region if any option's objects might be present but small. If after zooming objects are still absent, proceed to Step 1-V rule for all-absent.

\skillheading{Step 5 --- Note before answering}
Drop one note per option on the image at the region where that option's key object was found or sought. Note ordinal corresponds to option letter (1st note = A, 2nd note = B, 3rd = C, 4th = D). For absent objects, drop the note at the region where the object would be if it existed. The canvas colors these rings distinctly so you can visually confirm your letter mapping.

\skillheading{Step 6 --- Answer}
Read your VERDICT TABLE. Copy the FINAL LETTER to the answer field. That letter and only that letter goes in the answer field.

State the single letter only (or the exact number of letters the question format requires --- usually one). Never answer 'Left' or 'Right' as a standalone string. Never list all four option letters unless the question explicitly asks for all that apply.

\skillheading{Anti-patterns to avoid}
\begin{itemize}
\item Do NOT answer ABCD when only one answer is required, even if all options describe absent objects --- select the single most specifically incorrect option.

\item Do NOT skip the noun-grounding pre-pass (Step 1) or the verdict table (Step 1-V).

\item Do NOT claim options are missing --- they are always in the question text.

\item Do NOT answer 'Left' or 'Right' if the options are A/B/C/D.

\item Do NOT skip zooming on slide/panel grids or embedded screens.

\item Do NOT issue more zoom steps than needed once the answer is clear.

\item Do NOT let a letter appear in the answer field before the verdict table is complete.

\item Do NOT select a letter based on what 'sounds plausible in the real world' when the image shows a completely different scene --- always rank by visual contradiction to the actual scene.

\end{itemize}

\subsection{Visual configuration: VIEW.yaml}
\skillheading{Rendering parameters}
\begin{tcolorbox}[skillpanel]
\begin{Verbatim}[fontfamily=zi4,fontsize=\footnotesize,breaklines=true,breakanywhere=true,breaksymbolleft={},breaksymbolright={},breakindent=0pt,numbers=none]
allow_upscale: false
crop_long_edge_px: 1024
crop_margin: 0.02
grid: null
max_crops_per_step: 2
min_crop_frac: 0.0
overview_long_edge_px: 1024
pixel_expr: ''
select_expr: ''
union_crop_for_multi_target: true
view_ops: []
zoom_gate_expr: zoom_gain > 1.05 or region_frac < 0.12
\end{Verbatim}
\end{tcolorbox}
\skillheading{Canvas expression: \texttt{canvas\_expr}}
\begin{tcolorbox}[skillpanel]
\begin{Verbatim}[fontfamily=zi4,fontsize=\footnotesize,breaklines=true,breakanywhere=true,breaksymbolleft={},breaksymbolright={},breakindent=0pt,numbers=none]
opt_a_r = where(nearest_idx == 1, clip(red * 0.4 + 220, 0, 255), red)
opt_a_g = where(nearest_idx == 1, clip(green * 0.4 + 40, 0, 255), green)
opt_a_b = where(nearest_idx == 1, clip(blue * 0.4 + 40, 0, 255), blue)
opt_b_r = where(nearest_idx == 2, clip(opt_a_r * 0.4 + 20, 0, 255), opt_a_r)
opt_b_g = where(nearest_idx == 2, clip(opt_a_g * 0.4 + 180, 0, 255), opt_a_g)
opt_b_b = where(nearest_idx == 2, clip(opt_a_b * 0.4 + 180, 0, 255), opt_a_b)
opt_c_r = where(nearest_idx == 3, clip(opt_b_r * 0.4 + 40, 0, 255), opt_b_r)
opt_c_g = where(nearest_idx == 3, clip(opt_b_g * 0.4 + 80, 0, 255), opt_b_g)
opt_c_b = where(nearest_idx == 3, clip(opt_b_b * 0.4 + 220, 0, 255), opt_b_b)
opt_d_r = where(nearest_idx == 4, clip(opt_c_r * 0.4 + 220, 0, 255), opt_c_r)
opt_d_g = where(nearest_idx == 4, clip(opt_c_g * 0.4 + 200, 0, 255), opt_c_g)
opt_d_b = where(nearest_idx == 4, clip(opt_c_b * 0.4 + 20, 0, 255), opt_c_b)
ring_mask = where(dist < 0.055, where(dist > 0.030, 1, 0), 0)
base_r = where(ring_mask > 0, opt_d_r, where(dist < 0.030, clip(red * 0.5, 0, 255), red))
base_g = where(ring_mask > 0, opt_d_g, where(dist < 0.030, clip(green * 0.5, 0, 255), green))
base_b = where(ring_mask > 0, opt_d_b, where(dist < 0.030, clip(blue * 0.5, 0, 255), blue))
rgb(base_r, base_g, base_b)
\end{Verbatim}
\end{tcolorbox}

\Needspace{8\baselineskip}
\subsection{Interface: LAYOUT.yaml}
\skillheading{Request and history parameters}
\begin{tcolorbox}[skillpanel]
\begin{Verbatim}[fontfamily=zi4,fontsize=\footnotesize,breaklines=true,breakanywhere=true,breaksymbolleft={},breaksymbolright={},breakindent=0pt,numbers=none]
annotate_overview_boxes: false
context_ops: []
evict_policy: fifo
history_keep_last_n: 3
keep_overview: true
layout_expr: ''
max_images_per_request: 6
order: overview_first
\end{Verbatim}
\end{tcolorbox}
\skillheading{View contract: \texttt{view\_contract}}
Image 0 is the single composite source image containing all panels, sub-images, slides, or scene content referenced by the question. Notes burned into the image appear as colored rings: note 1 (option A) = red/orange ring, note 2 (option B) = teal/cyan ring, note 3 (option C) = blue ring, note 4 (option D) = yellow ring. Each ring marks the region where that option's key object was sought --- either found there, or the empty region where it would be if it existed. The ring color encodes option letter identity so the verdict table letter-to-region mapping can be visually confirmed.

\Needspace{9\baselineskip}
\skillheading{Answer contract: \texttt{answer\_contract}}
A single uppercase option letter (e.g. A, B, C, or D). The answer field must contain ONLY this one letter --- no other letters, no punctuation, no words. The letter must be copied from the FINAL LETTER line of the VERDICT TABLE written in the note field. For negation-framed questions the FINAL LETTER is the option marked ABSENT/INCORRECT. For affirmative questions the FINAL LETTER is the option marked APPLIES. Place all reasoning, the full verdict table, and evidence summaries in the note field only.

\skillheading{Protocol contract: \texttt{protocol\_contract}}
You are a visual reasoning agent. Each step you output exactly ONE JSON action block.

\textbf{Actions}

\begin{itemize}
\item \texttt{zoom}: \{"action":"zoom","regions":[\{"x0":...,"y0":...,"x1":...,"y1":...\}],"image":N,"note":"why"\} --- coordinates are plain decimals in [0,1] in image N's frame, x0\textless{}x1, y0\textless{}y1. Up to 2 regions per step. Zooming re-reads original file so real detail is recovered. Zooming inside a zoom is allowed.

\item \texttt{note}: \{"action":"note","notes":[\{"at":[x,y],"text":t,"image":N\}]\} --- persistent marker visible in later views and zooms.

\item \texttt{view}: \{"action":"view","units":[i,...]\} --- attach additional source units (pool size: \{pool\_size\}).

\item \texttt{compose}: \{"action":"compose","units":[...],"cols":c\} --- tile units into a labelled comparison sheet.

\item \texttt{answer}: \{"action":"answer","answer":"LETTER","note":"what you saw"\}

\end{itemize}
\textbf{Domain guidance}

\begin{enumerate}
\item Read the full image first. Identify panel count and question type before doing anything else.

\item Answer in one step if the content is unambiguous. Zoom when slide text, panel labels, or fine scene detail is needed.

\item The answer options are in the question text --- never say they are missing.

\item Always map the answer to the option LETTER (A/B/C/D). Never answer 'Left' or 'Right' as a standalone string.

\item Drop a \texttt{note} on the deciding evidence region before issuing \texttt{answer}.

\item Zoom budget: use only as many crops as needed. \{max\_crops\} crops maximum.

\end{enumerate}
\{ans\_header\} Example: \{ans\_example\}

\endgroup

\FloatBarrier
\section{Additional Related Work}
\label{app:additional-related-work}
Beyond skill evolution, multimodal large models enable diverse visual applications. Personalization research explores user-specific concept understanding and generation~\citep{an2024mc,an2025unictokens,yang2025smalllarge,zeng2026mira}, alongside retrieval and memory for long-term interaction and streaming video~\citep{xu2025jarvis,feng2026m2a,zheng2026pearl}. Visual applications encompass region-level image and video perception, advertising-video reasoning, dataset subpopulation analysis, and visually grounded agent skills~\citep{lin2025perceiveanythingrecognizeexplain,xu2026admir,luo2024llm,xu26}. Image and video editing agents use visual feedback or structured tool plans to guide edits~\citep{zeng2026mira,yu2026aurora}. Complementary work examines contextual visual generation and its evaluation~\citep{an2026genius}, alongside pixel-space diffusion representations and approaches that optimize diffusion models using rewards~\citep{yan2026pixelspace,yan2025entropyadaptive,yan2026doless}.

\end{document}